\documentclass[10pt,twocolumn,letterpaper]{article}

\usepackage[pagenumbers]{cvpr} 

\usepackage{makecell}

\definecolor{tabfirst}{rgb}{1, 0.7, 0.7} 
\definecolor{tabsecond}{rgb}{1, 0.85, 0.7} 
\definecolor{tabthird}{rgb}{1, 1, 0.7} 

\definecolor{Red}{HTML}{FFB3B3} 
\definecolor{Orange}{HTML}{FFD9B3}
\definecolor{Yellow}{HTML}{FFFFB3}

\definecolor{cvprblue}{rgb}{0.21,0.49,0.74}
\usepackage[pagebackref,breaklinks,colorlinks,allcolors=cvprblue]{hyperref}

\usepackage{tabularx}
\usepackage{amssymb} 

\usepackage{colortbl} 
\usepackage{xcolor}

\usepackage{graphicx}
\usepackage{adjustbox}

\usepackage[normalem]{ulem}

\usepackage{amsmath}
\usepackage{amssymb}
\usepackage[ruled,vlined]{algorithm2e}

\def\paperID{***} 
\def\confName{CVPR}
\def\confYear{2026}

\newcommand{\methodname}{Turbo-GS\xspace}
\title{\methodname: Accelerating 3D Gaussian Fitting for High-Resolution Radiance Fields}

\author{Ankit Dhiman$^{*\,1,4}$ \quad
Tao Lu\thanks{} $^{\,2}$ \quad
 R Srinath$^{*\,1}$ \quad
 Emre Arslan$^2$  \quad
 Angela Xing$^2$ \\
 Yuanbo Xiangli$^3$ \quad
 Venkatesh Babu Radhakrishnan$^1$  \quad
 Srinath Sridhar$^2$ \\
  \quad
$^1$Indian Institute of Science, Bangalore $^2$Brown University\\ $^3$Cornell University \quad $^4$Samsung R\&D Institute India - Bangalore \\
\href{https://ivl.cs.brown.edu/research/turbo-gs}{https://ivl.cs.brown.edu/research/turbo-gs}
}

\begin{document}

\twocolumn[{%
	\renewcommand
	\twocolumn[1][]{#1}%
	\maketitle
	\begin{center}
		\centering
		\vspace{-0.3in}
            \includegraphics[width=0.99\textwidth]{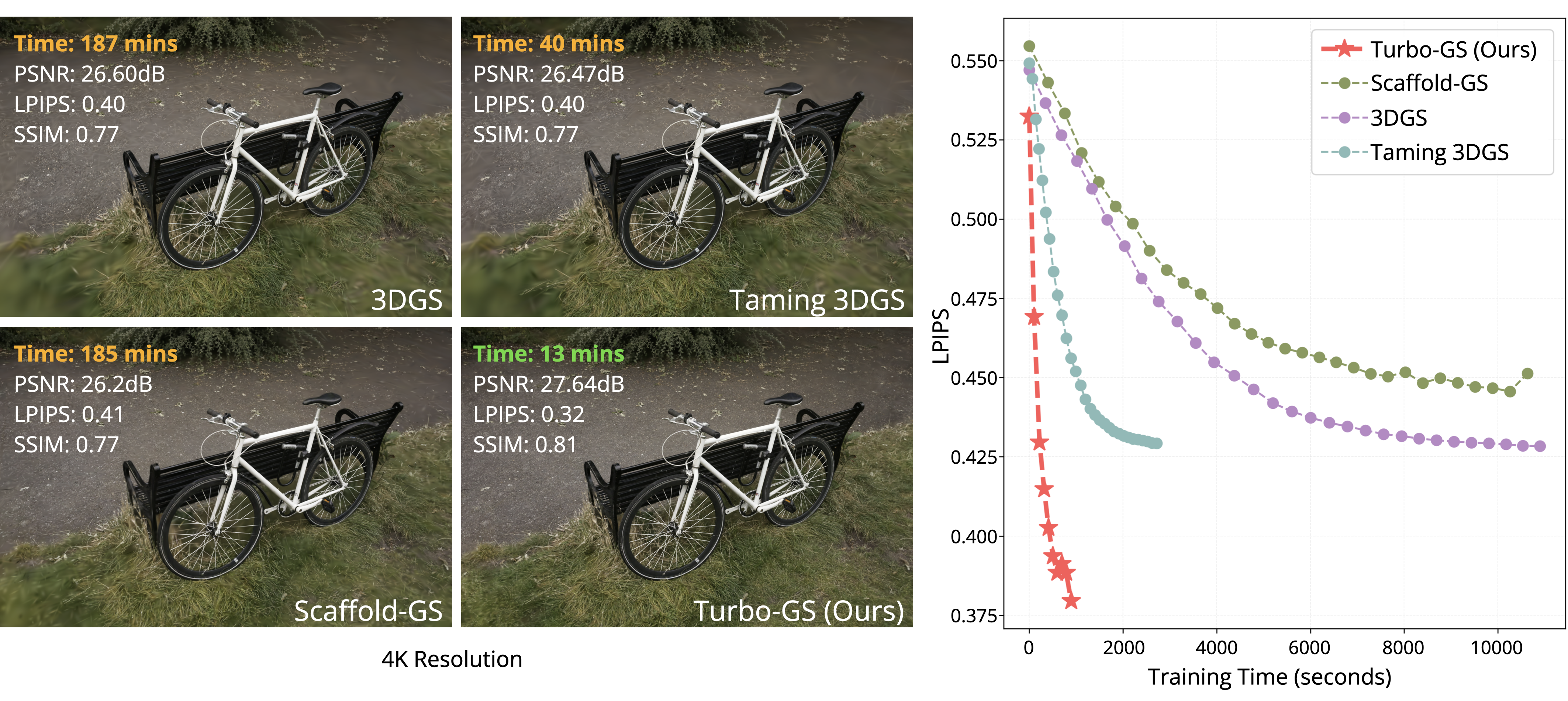}
        \vspace{-15pt}
		\captionof{figure}{\small
           \textbf{\methodname} accelerates 3DGS fitting significantly while preserving rendering quality. It proposes an efficient densification strategy and innovative dilated rendering that allow training on 4K images in minutes, significantly outperforming baseline methods.
           Notably, \methodname converges on the 4K bicycle scene in just \textbf{13 minutes}—over \textbf{3}$\times$ faster than Taming 3DGS (40 minutes), \textbf{14}$\times$ faster than 3DGS (187 minutes) and Scaffold-GS (185 minutes).
		}
		\label{fig:teaser}
	\end{center}
}]
\let\thefootnote\relax\footnotetext{*Equal Contribution.}

\maketitle

\begin{abstract}

Novel-view synthesis plays a crucial role in computer vision with applications in 3D reconstruction, mixed reality, and robotics. Recent approaches, such as 3D Gaussian Splatting (3DGS), have emerged as state-of-the-art solutions, offering high-quality novel view synthesis in real time. However, training 3DGS models remains slow, particularly for high-resolution images, often requiring hours to fit a scene with 200 views.
In this work, we aim to accelerate the fitting process by reducing computational overhead and improving learning efficiency. Specifically, we introduce a dilated rendering technique that renders only a subset of pixels instead of the full image, significantly reducing computational costs. To enhance learning efficiency, we develop a convergence-aware budget control mechanism that balances the addition of new Gaussians with the optimization of existing ones. Additionally, to improve densification efficiency and prevent gradient vanishing, we incorporate both positional and appearance errors to improve the effectiveness of densification.
With these improvements, we achieve fast 4K-resolution fitting while maintaining, or even improving, novel view rendering quality. Extensive experiments demonstrate that our method achieves significantly faster optimization than existing approaches while preserving high rendering fidelity. 
\end{abstract}
\vspace{-0.2in}    
\addtocontents{toc}{\protect\setcounter{tocdepth}{-2}}
\section{Introduction} 
\label{sec:intro}
Building the radiance field~\cite{Mildenhall2020NeRF} of a scene from multiple posed RGB images has recently become an important problem in computer vision given numerous applications in photorealistic novel view synthesis (NVS)~\cite{mip360,instantngp,plenoxels,3dgs,scaffoldgs}, 3D reconstruction~\cite{yariv2023bakedsdf,guedon2024sugar}, mixed reality~\cite{deng2022fov}, and robotics~\cite{ze2023gnfactor, dai2023graspnerf}.
While radiance fields were initially represented \emph{implicitly} using neural networks~\cite{Mildenhall2020NeRF,mip360}, primitive-based \emph{explicit} methods have become more popular~\cite{plenoxels,plenoctrees,tensorf}.
In particular, 3D Gaussian Splatting (3DGS)~\cite{3dgs}
has become the method of choice for representing radiance fields.
3DGS generates high-quality novel views by using a differentiable renderer based on Gaussian splat rasterization, achieving real-time rendering rates for 1080p (1K) images on a GPU.

Despite the progress in the NVS quality and rendering times, \emph{fitting (or optimizing)} a high-quality radiance field from posed images remains slow.
For example, using 3DGS to fit a static scene with 200 camera views at 1K resolution take about 30 minutes and 4K resolution about hours, depending on the scene.
This poses a significant challenge to the widespread adoption of radiance fields in broader problems, including modeling dynamic scenes and semantics.

Some prior works have recognized this problem and proposed solutions.
For example, learning-based methods~\cite{instantsplat,gslrm2024,longlrm, lvsm, mvsplat,noposplat} estimate intermediate Gaussian initializations or directly estimate the final Gaussian positions in a feed-forward manner, enabling reconstruction in a few seconds.
However, these methods are limited to a fixed number of input views and low resolution, let alone handling multiple high-resolution 4K images. And their generalization abilities are not well studied.
Among learning-free methods that can handle an arbitrary number of views, 
the focus has been on (1) building highly-optimized CUDA implementations, like accelerating the backward~\cite{mallick2024taming}, accurate tile-Gaussian pair construction~\cite{adrgaussian, stopthepop, speedygs} and load balance~\cite{adrgaussian}; (2) lightweight encodings to quantize Gaussian attributes~\cite{Girish2023EAGLESEA}; (3) strategies to reduce the number of gaussian points~\cite{fang2024mini}; or (4) replacing the optimizer with second-order methods for quicker convergence~\cite{Hllein20243DGSLMFG}.
However, their approach neglects the analysis on high-resolution scenes, which are crucial for immersive experiences.

\begin{figure}[!t]
    \centering
    \includegraphics[width=0.96\linewidth]{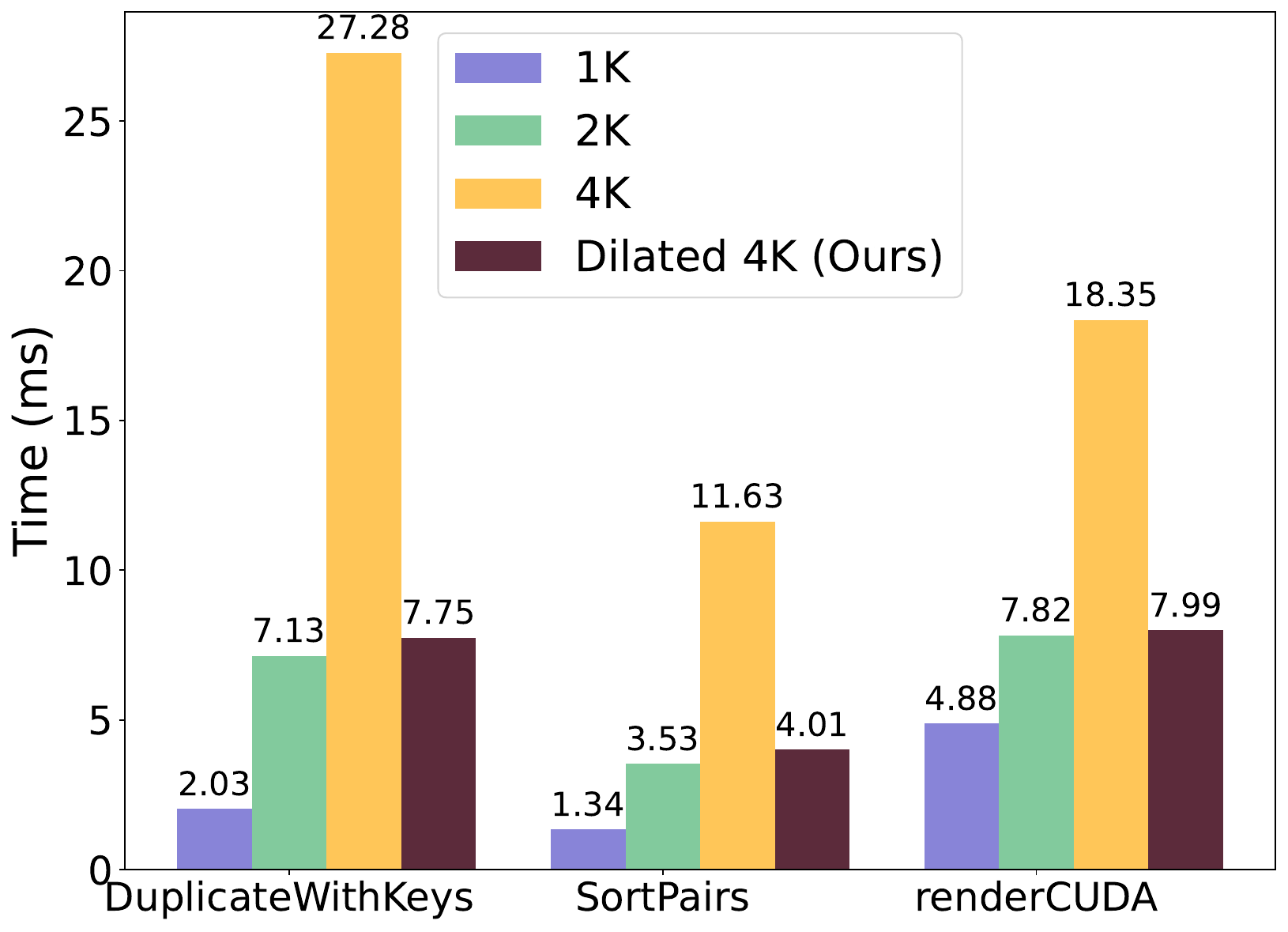}
    \caption{\textbf{Runtime breakdown}. We rasterize different resolution images using the same Gaussian checkpoint and profile the three most time-consuming components. We observed that the overhead of DuplicateWithKeys ( for associating Gaussian to tiles) and SortPairs (for obtaining depth order) increase proportionally to the image size, and is even more expensive than the renderCUDA (for alpha-blending) in high-resolution task.}
    \label{fig:profile}
    \vspace{-5mm}
\end{figure}

In this paper, we focus on accelerating the per-scene optimization of 3DGS in high-resolution setting \emph{without using any learning-based prior and without sacrificing quality}. To this end, we introduce \textbf{\methodname}, a method that fundamentally improves the efficiency of high-resolution 3DGS training, while maintaining or exceeding state-of-the-art rendering quality.
Different from previous methods, we emphasize on high-resolution~(\eg 4K resolution) tasks because the training overhead issue is more significant (See in ~\cref{fig:profile}). We provide detailed explanation in~\cref{sec:dilated_rendering}.
The key idea is to (1) reduce the computational overhead by minimizing the footprint and (2) enhance the optimization efficiency of each iteration.
Previous methods have extensively analyzed and optimized CUDA modules from the perspectives of computation, scheduling, and memory access. In contrast, we shift our focus to the supervised training process, questioning whether each resource unit used for optimization is necessary.
Our key insight is that, since 3DGS employs tile-based rendering, each Gaussian can influence multiple pixels. Consequently, dense pixel-wise supervision often introduces redundancy, particularly at higher resolutions. To mitigate this, we propose rendering only a subset of pixels instead of the full image for supervision. Based on this principle, we design a coordinate mapping process that seamlessly enables dilated rendering without inducing idle GPU threads. This approach can be integrated with existing efficiency improvements in CUDA rasterizer.
To further enhance optimization efficiency, we introduce a position-appearance guidance strategy for densification, preventing vanishing gradients in textureless regions. Additionally, to balance densification and Gaussian fitting quality, we propose a convergence-aware budgeting mechanism for adding new Gaussians, ensuring a more efficient and stable training process.

With these contributions, our approach reduces 4K scene reconstruction time from several hours to just 10 minutes.
We conduct extensive experiments to validate our design choices and compare our approach with previous methods.
To sum up our contributions:
\begin{itemize}
    \item We propose \textbf{\methodname}, a method for fitting 3DGS that is several times faster than existing methods. Our key idea is to maximize effective optimization.
    \item We implement a highly-efficient dilated rendering pipeline, which could be integrated into different 3DGS variants.
    \item We optimize the densification process by dynamically adjusting the densification rate and enhancing the guidance.
\end{itemize}

\section{Related Work} 
\label{sec:related}

\subsection{Novel View Synthesis}
Novel-view synthesis has gained significant traction in recent years, with Neural Radiance Fields (NeRF)~\cite{Mildenhall2020NeRF} emerging as a standout technique for generating highly photorealistic images. 
NeRF achieves this by leveraging volume rendering to optimize multi-layer perceptron (MLP) weights, 
but the original approach is computationally expensive and requires several hours or even days for training. 
To address this, subsequent works have integrated NeRF with explicit representations like voxel and feature grids~\cite{Sun2021DirectVG,Karnewar2022ReLUFT,plenoxels,Xu2023GridguidedNR}, hash grids~\cite{instantngp}, and point-based methods~\cite{Xu2022PointNeRFPN} to dramatically accelerate training times.

3D Gaussian Splatting (3DGS)~\cite{3dgs} extends these developments by modeling scenes as a collection of 3D Gaussians that are projected as 2D splats and combined using alpha blending to form pixel colors. 
This technique has gained popularity for enabling high-quality, real-time rendering. 
Scaffold-GS~\cite{scaffoldgs} further enhanced 3DGS by introducing a hierarchical structure that aligns anchors
with scene geometry.
They introduced a multi-resolution error-based densification strategy that further enhances the robustness of the adaptive control of Gaussians, improving both rendering quality and memory efficiency.
Meanwhile, numerous works have focused on various enhancements to 3DGS, including improved rendering quality~\cite{Yu2023MipSplattingA3,Huang20242DGS}, faster rendering~\cite{Fan2023LightGaussianU3,Girish2023EAGLESEA}, better surface reconstruction accuracy~\cite{Yu2024GSDF3M,guedon2024sugar}, memory optimization~\cite{Ren2024OctreeGSTC,Papantonakis2024ReducingTM}, and the ability to handle large-scale scenes~\cite{Kerbl2024AH3}.
While 3DGS delivers high-quality results with extremely fast rendering, it faces challenges such as unpredictable storage requirements and variable fitting durations, which can hinder its effectiveness in downstream applications.

\subsection{Accelerating Gaussian Splatting Fitting}
Recognizing the limitation in fitting speed, some recent work has focused on accelerating fitting and improving efficiency by enhancing initialization methods~\cite{Jung2024RelaxingAI,Paliwal2024CoherentGSSN,Kheradmand20243DGS,Foroutan2024EvaluatingAT}, optimizing Gaussian budget (number of Gaussians) allocation~\cite{fang2024mini,mallick2024taming}, and reducing overall training time~\cite{Girish2023EAGLESEA,Navaneet2023CompGSSA,Hllein20243DGSLMFG}. 

In terms of initialization, methods such as Rain-GS~\cite{Jung2024RelaxingAI} and Gaussian Splatting as MCMC~\cite{Kheradmand20243DGS} enable effective model fitting from sub-optimal, or randomly initialized point clouds, expanding the robustness of 3DGS in diverse settings. 
CoherentGS~\cite{Paliwal2024CoherentGSSN} leverages monocular depth and dense flow correspondences to provide a well-defined set of initial 3D Gaussians, while studies exploring alternatives to SfM-based initialization~\cite{Foroutan2024EvaluatingAT} have shown that combining improved random initialization with NeRF’s structural guidance achieves, or even surpasses, the quality of COLMAP initialization on large-scale, challenging scenes.

In managing the Gaussian budget, Mini-Splatting~\cite{fang2024mini} identifies spatial inefficiencies within Gaussian distributions and introduces strategies like blur splitting, depth reinitialization, and intersection-preserving sampling to address these redundancies.
Taming 3DGS~\cite{mallick2024taming} extends this approach by adapting Gaussian distribution to specific use cases while prioritizing perceptual quality through a flexible, score-based framework.
Efforts to reduce training time, including EAGLES~\cite{Girish2023EAGLESEA} and Compact3D~\cite{Navaneet2023CompGSSA}, use quantization to streamline storage and computation, while 3DGS-LM~\cite{Hllein20243DGSLMFG} replaces the Adam optimizer with the LM optimizer to accelerate convergence.
Other works improve the optimization runtime by improving the implementation of the underlying differentiable rasterizer~\cite{Durvasula2023DISTWARFD,Feng2024FlashGSE3,Ye2024gsplatAO}.

A line of research has also focused on sparse-view reconstruction using a data-driven, feed-forward approach, directly generating Gaussians in a single forward pass~\cite{Charatan2023PixelSplat3G,Chen2024LaRaEL,gslrm2024,mvsplat,instantsplat}. 
In this work, we do not rely on priors from large foundation models, as they have limited capability to accommodate densely captured views and generally cannot handle high-resolution scene modeling.

\begin{figure*}[thbp]
    \centering
    \includegraphics[width=0.95\linewidth]{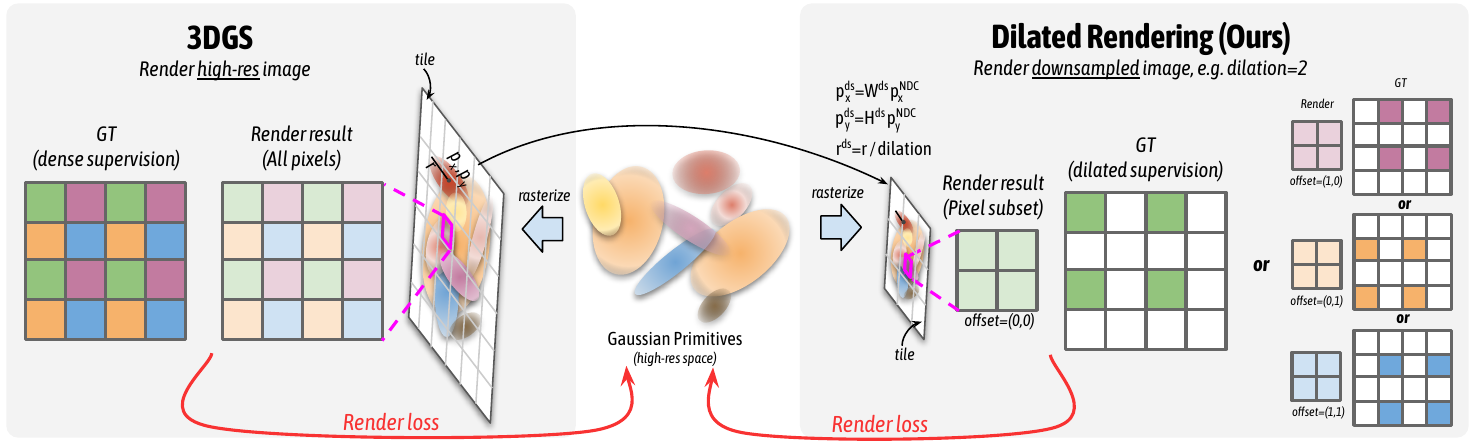}
    \caption{\textbf{Dilated Rendering Pipeline.} 3DGS~\cite{3dgs} renders full high-resolution images for supervision, which is computationally expensive. We propose a dilated rendering approach, where only a subset of pixels, sampled in a dilated pattern, is rendered. Specifically, we generate tile-Gaussian pairs in the downsampled image space by mapping image-plane Gaussian centers \((p_x, p_y)\) and search radius \(r\) from the high-resolution space to \((p^{ds}_x, p^{ds}_y)\) and \(r^{ds}\) in the downsampled space. Alpha-blending is performed in the high-resolution space with supervision from corresponding high-resolution pixels. For illustration, we use a dilation factor of 2, where pixel sampling is controlled by an offset \( O=\{(i,j)\}, i,j \in \{0,1\} \), and at each iteration, a random offset is selected for rendering a low-resolution image.
   }
    \label{fig:dilated_rendering} 
\end{figure*}
\section{Preliminaries}
\label{sec:preliminary}
In this section, we present a brief overview of our baseline methods: the explicit 
 method 3DGS~\cite{3dgs} and explicit-implicit method Scaffold-GS~\cite{scaffoldgs}.
3DGS models scene geometry as a set of 3D Gaussian primitives. Each 3D Gaussian is formulated as:
\begin{equation}
    G(x)=e^{-\frac{1}{2}(x-\mu)^T \Sigma^{-1}(x-\mu)},
\end{equation}
where $\mu$ is the center position and $\Sigma$ is an anisotropic covariance matrix. The covariance matrix is defined by a rotation matrix $R$ and scaling matrix $S$ as $RSS^TR^T$. Additionally, each Gaussian has an opacity parameter $\sigma$ and spherical harmonics (SH) coefficients to model view-dependent color. To render a given viewpoint, the Gaussians are projected as 2D splats, sorted by depth, and combined with $\alpha$-blending using a tile-based rasterizer. The color $C$ of each pixel is defined by: 
\begin{equation}
    C\left(x^{\prime}\right)=\sum_{i}c_i \sigma_i \prod_{j=1}^{i-1}\left(1-\sigma_j\right), \quad
    \sigma_i=\alpha_i G_i^{\prime}\left(x^{\prime}\right), 
\end{equation}
where $x^{\prime}$ is the queried pixel, $c_i$ is the color of the $i$-th Gaussian, $\alpha_i$ is the learned opacity for the $i$-th Gaussian, and $G_i^{\prime}\left(x^{\prime}\right)$ is the 2D projection of the Gaussian at pixel $x^{\prime}$. 3DGS begins with initialization from sparse SfM point cloud and after every interval, Gaussian primitives are pruned, split and cloned based on the accumulated gradient.  
Scaffold-GS extends 3DGS to provide more structure to the scene by introducing anchor points and spawning $k$ neural Gaussians which are predicted through MLP for each anchor point.

\section{\methodname: Accelerated 3D Gaussian Fitting} 
\label{sec:method}

We aim to develop a 3DGS optimization framework in particular for high-resolution scenarios~(\eg 4K).
The key idea is to: (1) Reduce computational overhead by minimizing the footprint of each optimization step using a novel \emph{dilated rendering} technique (\cref{sec:dilated_rendering}), while preserving high-resolution rendering quality; (2) Enhance optimization efficiency through a \emph{convergence-aware training schedule} (\cref{sec:convergence_aware_training_schedule}), maximizing performance gains at each step; (3) Improve densification effectiveness and efficiency with a \emph{position-appearance-based strategy} (\cref{sec:pos_appearance_densification}), ensuring more adaptive and robust updates.

\subsection{Dilated Rendering}
\label{sec:dilated_rendering}

Directly fitting 3D Gaussian Splatting (3DGS) on high-resolution inputs suffers from intensive computation and high memory demand.
To illustrate this, we profiled the execution time of tile-Gaussian generator~(\ie \texttt{duplicateWithKeys}), sorting~(\ie \texttt{sortPairs}) and alpha-blending~(\ie \texttt{renderCUDA}) during rasterization of the same set of 3D Gaussians at different image resolutions. Results are shown in~\cref{fig:profile}.
As image resolution increases, the time required for tile-Gaussian generation and sorting scales proportionally with the number of pixels, while alpha-blending also becomes more computationally expensive. Additionally, higher resolutions lead to a greater number of tile-Gaussian pairs, resulting in increased memory usage too.

A straightforward strategy is to render a subset of pixels instead of the full image for loss computation (similar to NeRF~\cite{Mildenhall2020NeRF} which forwards a set of sampled rays). However, in 3D Gaussian Splatting (3DGS), the tile-based rendering process and the complex projection of Gaussian functions make it inherently challenging to decouple the computation of individual rays.

To this end, we propose the novel dilated rendering pipeline for accelerating 3D Gaussian Splatting fitting. We use a chessboard sampling pattern~(\ie dilated) to obtain a subset of pixels from the original image, forming a downsampled image, as shown in~\cref{fig:dilated_rendering}, controlled by three parameters: a dilation size $d$ and two offsets $(o_x, o_y)$ for horizontal and vertical directions. 
The core of our design involves: (1) generating a tile-Gaussian pair from the downsampled image, and (2) performing alpha-blending in the original (high-resolution) image space.

Specifically, to derive tile-Gaussian pairs from the downsampled image, we need to compute the Gaussian centers and determine a search radius on the downsampled image plane.
To do so,
we project a Gaussian's 3D position \(p\) into a resolution‐independent normalized device coordinate (NDC) space, yielding \(p^{\mathrm{NDC}}\); 
then map it onto the downsampled image plane via:
\begin{equation}
p_x^{\mathrm{ds}} = W^{\mathrm{ds}}\, p_x^{\mathrm{NDC}}, \quad
p_y^{\mathrm{ds}} = H^{\mathrm{ds}}\, p_y^{\mathrm{NDC}},
\end{equation}
where \(W^{ds}\) and \(H^{ds}\) denote the width and height of the downsampled image.
Next, we compute each Gaussian's 2D covariance matrix
using the original focal length; 
and update the covariance with a low-pass filter to ensure covering at least one pixel following~\cite{3dgs}.
From this covariance, we derive the 3-\(\sigma\) radius \(r\) corresponds to the high-resolution image space. 
%
We then obtain the radius $r^{\mathrm{ds}}$ on the downsampled image by:
\begin{equation}
r^{\mathrm{ds}} = r/d,
\end{equation}
\noindent
from which the occupied tiles are computed accordingly. 

During alpha-blending, each pixel coordinate from the downsampled image space is mapped back to the original high-resolution image space using the following transformations:

\begin{equation}
\begin{aligned}
p_x &= d \, p_x^{\mathrm{ds}} + \frac{1}{2}\,(d - 1),\\[1mm]
p_y &= d \, p_y^{\mathrm{ds}} + \frac{1}{2}\,(d - 1).
\end{aligned}
\end{equation}

\begin{figure}[t]
  \centering
   \includegraphics[width=1\linewidth]{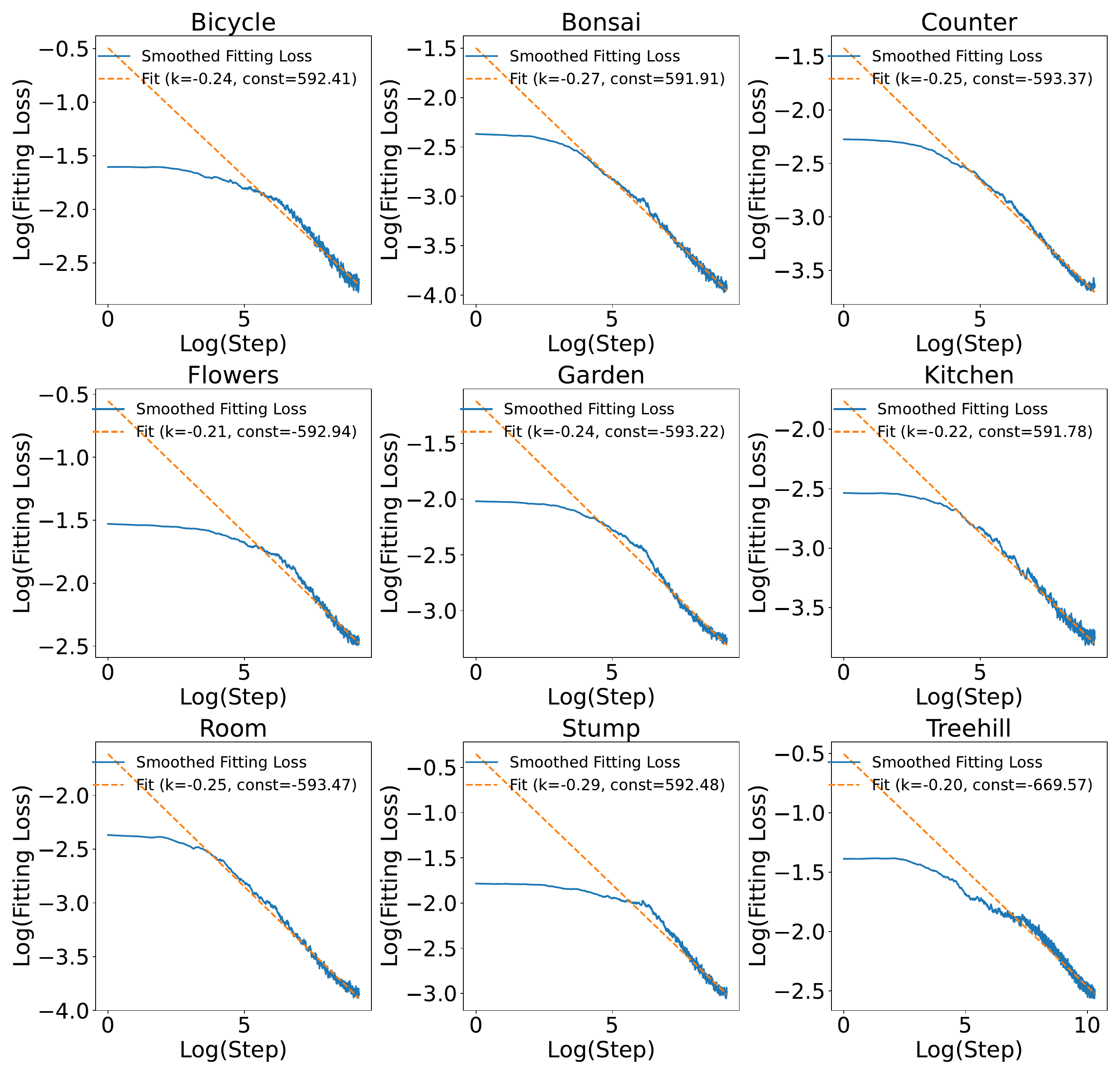}
    \vspace{-20pt}
   \caption{\textbf{Loss analysis with power function fitting.} For all scenes, the $log$(loss) is linear to the $log$(iterations) after the initial stage. Thus, the relation between iteration and convergence follows a power function. We design a power-law-based adaptive budget schedule based on these insights.
   } 
   \vspace{-10pt}
   \label{fig:powerfit}
\end{figure}

\noindent 
This design enables high-resolution rendering with significantly lower computational cost, as GPU computation, scheduling, and memory access remain consistent with those in the downsampled image space. Meanwhile, alpha-blending in the high-resolution space preserves rendering quality without degradation.

Note that, to adapt to the GPU architecture, the tile size is a scale‐independent fixed number of $16\times16$. Therefore, a tile in the low‐resolution space covers an area equivalent to \(d^2\) tiles in the high‐resolution space. The difference in tiles configuration will cause a disturbance to the rendering result. To evaluate the error introduced by this misalignment, we conducted comparative validation. Given a trained Gaussian checkpoint, we first render high-resolution images with the original 3DGS rasterizer, denoted as $I$. Then for each view we render 4 low-resolution images with $d=2, (o_x,o_y)=\{(0,0),(1,0),(0,1), (1,1)\}$ and merge the 4 into a high-resolution image, denoted as $I'$. Evaluation show that the merged high-resolution images achieve extremely high level of fidelity, with a PSNR($I$, $I'$) exceeding 70dB. 

Based on this implementation, we train with a dilation factor $d=2$ and the offset set $O=\{(i,j)\}, i,j \in \{0,1\}$. For each iteration we randomly sample one offset to render a low-resolution image for supervision.

\subsection{Convergence-aware Training Schedule}
\label{sec:convergence_aware_training_schedule}

Denoting the initial number of Gaussian points as $N$, we manually set a maximum number $M$. The challenge then is how the number of Gaussians grows from $N$ to $M$. A proper budget mechanism to regulate newly added Gaussians helps control the growth rate of Gaussian and maximize improvement for reconstruction in each iteration. 

To create such a budget, we take a closer look at how the reconstruction process converges.
We calculated the logarithm of the training loss and iterations in~\cref{fig:powerfit},
and observed that after the initial stage, the $log(\text{loss})$ is almost perfectly linear to the $log(\text{iterations})$ in every scene (demonstrated on MipNeRF360~\cite{mip360}).
This means that the relation between fitting iterations and model convergence follows a power function, which can be ascribed to the use of MSE loss~\cite{liang2024scaling} and L1 loss.
Based on this finding, we design a power-law-based adaptive budget schedule.

\begin{figure}[!t]
  \centering
   \includegraphics[width=1\linewidth]{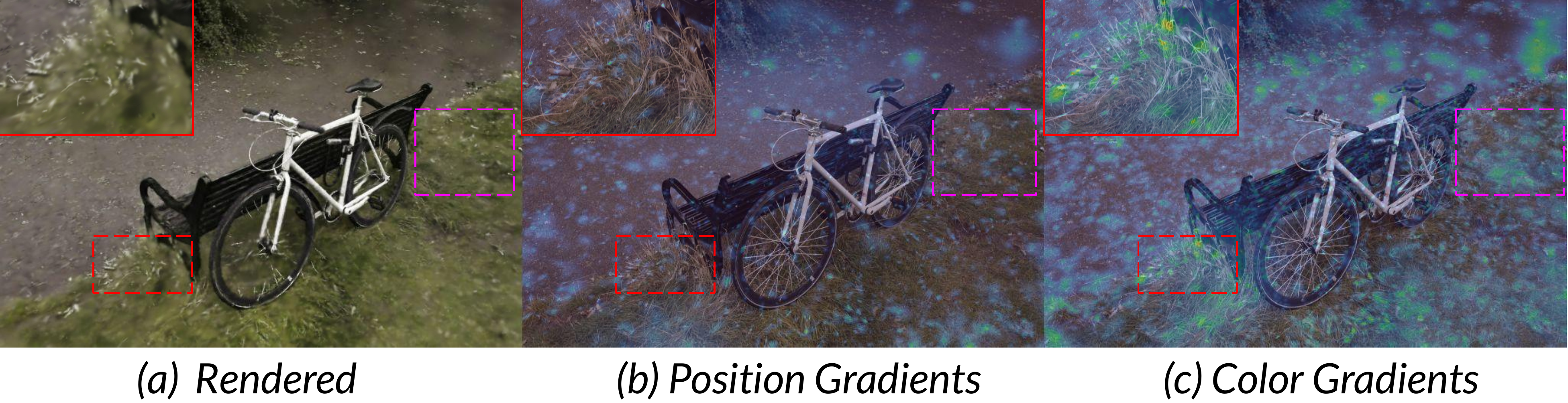}
    \vspace{-20pt}
   \caption{\textbf{Gradient Visualization.} We perform alpha-blending of gradients for each Gaussian into image plane. \textbf{(a)} Rendered Image. We observe that: (b) Position Gradients focus only on certain regions in the scene, while (c) Color Gradients provide cues from overall regions, especially blurry regions. Darker regions indicate smaller gradients and brighter regions represent larger gradients.
   }
   \vspace{-10pt}
   \label{fig:gradient_viz}
\end{figure}

\begin{table*}[t]
\caption{Quantitative comparison with Gaussian Splatting baselines for 4K scenes. We compare PSNR, SSIM, and LPIPS for quality. For resource efficiency, we report training time, memory and, where applicable and peak number (Peak \#G) of Gaussians or Anchors (for Scaffold-GS and its variants) used. '-accel' means accelerated with optimized CUDA implementation. For Scaffold-GS-accel, it also means training for 10k steps.}
\vspace{-5pt}
\begin{adjustbox}{width=\linewidth}
\begin{tabular}{l|c|c|c|c|c|c|c|c|c|c|c|c|}
\multicolumn{1}{c}{} & \multicolumn{6}{c|}{MipNeRF-360 4K~\cite{mip360}} & \multicolumn{6}{c}{EyefulTower 4K~\cite{eyefultower}} \\ \cline{2-13} 
                             & \multicolumn{1}{c|}{SSIM$\uparrow$}           & \multicolumn{1}{c|}{PSNR$\uparrow$}           & \multicolumn{1}{c|}{LPIPS$\downarrow$}          & \multicolumn{1}{c|}{\begin{tabular}[c]{@{}c@{}}Train\\ time $\downarrow$\end{tabular}} & \multicolumn{1}{c|}{\begin{tabular}[c]{@{}c@{}}Memory $\downarrow$\end{tabular}} & \multicolumn{1}{c|}{\begin{tabular}[c]{@{}c@{}}Peak \\ \#G $\downarrow$\end{tabular}} & \multicolumn{1}{c|}{SSIM$\uparrow$}          & \multicolumn{1}{c|}{PSNR$\uparrow$}           & \multicolumn{1}{c|}{LPIPS$\downarrow$}          & \multicolumn{1}{c|}{\begin{tabular}[c]{@{}c@{}}Train\\ time $\downarrow$\end{tabular}} & \multicolumn{1}{c|}{\begin{tabular}[c]{@{}c@{}}Memory $\downarrow$\end{tabular}} & \multicolumn{1}{c|}{\begin{tabular}[c]{@{}c@{}}Peak \\ \#G $\downarrow$\end{tabular}} \\
 \hline  
3DGS~\cite{3dgs}     & \cellcolor{Yellow}0.797 & \cellcolor{Yellow}26.75 & 0.410 & 113~m & 412~MB & 1.74~M & \cellcolor{Red}0.920 & \cellcolor{Red}31.18 & \cellcolor{Yellow}0.296 & 200m & 328~MB & \cellcolor{Orange}1.38~M \\
3DGS-accel~\cite{3dgs}     &0.791 & 26.52 & 0.364 & 40~m & 383~MB & \cellcolor{Yellow}1.62~M & \cellcolor{Yellow}0.918 & 30.89 & 0.302 & 45m & 323~MB & \cellcolor{Yellow}1.37~M \\
Turbo-3DGS (ours)     & \cellcolor{Red}0.805 & \cellcolor{Orange}26.80& \cellcolor{Orange}0.327& \cellcolor{Yellow}24~m & 680~MB & 2.80~M & \cellcolor{Orange}0.919 & \cellcolor{Orange}31.07 & \cellcolor{Orange}0.294 & \cellcolor{Yellow}27m & 553\,MB & 2.34~M \\\hline
Scaffold-GS~\cite{scaffoldgs}      &0.794  & \cellcolor{Red}26.84 & \cellcolor{Yellow}0.359 & 143~m & \cellcolor{Orange}154~MB & \cellcolor{Orange}1.33~M & \cellcolor{Red}0.920  & \cellcolor{Yellow}30.93 & \cellcolor{Red}0.292 & 213m & \cellcolor{Red}123~MB & \cellcolor{Red}1.06~M \\
Scaffold-GS-accel~\cite{scaffoldgs}      &0.784  & 25.81 & 0.373 & \cellcolor{Orange}12~m & \cellcolor{Red}72~MB & \cellcolor{Red}0.62~M & 0.903 & 27.53 & 0.330 & \cellcolor{Orange}14m & \cellcolor{Orange}191~MB & 1.63~M \\

Turbo-Scaffold-GS (ours)          & \cellcolor{Orange}0.800 & 26.65 & \cellcolor{Red}0.327 & \cellcolor{Red}9m  & \cellcolor{Yellow}269~MB & 2.33M & 0.915 & 30.21 & 0.305 & \cellcolor{Red}9m & \cellcolor{Yellow}293~MB & 2.52~M \\
\hline
\end{tabular}
\end{adjustbox}
\vspace{-10pt}
\label{tab:quant_table_4k}
\end{table*}

Specifically, in a power-law dominated system, the loss is expected to follow a power function. Therefore, our goal is to check whether the newly added Gaussian points have caused the convergence process to deviate from the power law. Starting with an initial power exponent \(\alpha_{\text{base}}\), we record the loss at each iteration after 100 warm-up steps. Periodically, we fit a historical power exponent \(\alpha_{\text{history}}\) using all recorded losses smoothed by exponential moving average. We then evaluate the loss over the most recent $k$ iterations to compute a local power exponent \(\alpha_{\text{recent}}\). The difference \(\epsilon = \alpha_{\text{recent}} - \alpha_{\text{history}}\) is used to adjust the power. The updated power is given by:
\begin{align}
    \alpha = \alpha_\text{base}+\lambda\cdot\tanh(\epsilon). 
    \label{eq:lambda_equation}
\end{align}
For simplicity, we set the $\alpha_\text{base}$ as the average value from historical $\alpha_\text{base}$ in our experiments, and we set $\lambda=0.5$. Then, the budget $B(t)$ for iteration $t$ is given by
\begin{align}
    B(t)=N + \frac{t^{\alpha} - 1}{100^\alpha - 1} * (M - N). 
    \label{eq:budget_schedule}
\end{align}

\noindent
This balances the growing speed of new Gaussians and the optimization speed of old Gaussians. If the convergence speed is lower than the expectation from power law, we then slow down the densification, and vice versa.

\subsection{Position-Appearance Based Densification}
\label{sec:pos_appearance_densification}

Previous works~\cite{scaffoldgs,3dgs} densify Gaussians in areas with strong positional gradients.
However, we observe that texture-less area usually have fewer Gaussian points and blur rendering effect. This is because in such regions, small changes in a Gaussian's position result in negligible error variations, leading to near-zero position gradient magnitude. A similar observation was also made in ~\cite{Bul2024RevisingDI}.
Inspecting gradients from different Gaussian attributes (visualized in~\cref{fig:gradient_viz}), we found that while positional gradients are limited in textureless background regions (e.g., grass), color gradients provide more informative signals. Therefore, we propose a straightforward remedy, which is to perform densification based on both color and position gradients.

Specifically, we threshold the position and appearance gradients with $\tau_{\text{position}}$ and $\tau_{\text{color}}$ respectively to determine whether densification is needed.
We observe that the color gradient has a smaller numerical range, so we set $\tau_{\text{color}}=0.01*\tau_{\text{position}}$.
Then, we follow baseline's design to add points with multi-scale voxels.
Since appearance-based densification tends to overfit the training view, we only activate it with a probability of 20\% to supplement the position-based densification.
We find that the color branch improves reconstruction even if it is not frequently activated.

\noindent
\textbf{Sensitivity-Based Pruning.} 
While dilated rendering skips a subset of pixels during training, we observe that this approximation does not impact the quality of Gaussian primitives produced during densification stage. To verify this, we inspect the sensitivity score introduced in Speedy-Splat~\cite{speedygs}, and find that the behavior remains consistent between models trained with and without dilated rendering. This consistency indicates that the primitives selected for densification are reliable. Building on this insight, we apply sensitivity-based pruning at a $60\%$ threshold, effectively removing redundant primitives while preserving the fidelity of the generated novel views.
\section{Experiments} 
\label{sec:exp}

We present implementation details of the proposed method in~\cref{subsec:implementation_details}.
We discuss the benchmark datasets and baseline methods in~\cref{subsec:dataset}. We present qualitative and quantitative results in~\cref{subsec:results}.
More details and results are available in the supplementary material.

\subsection{Implementation Details}
\label{subsec:implementation_details}

We build \methodname upon 3DGS~\cite{3dgs} and Scaffold-GS~\cite{scaffoldgs}, further enhancing them with the optimized CUDA kernel from Taming 3DGS~\cite{mallick2024taming} for efficient backward propagation.
For 3DGS-based experiments, we train models for 30k iterations. For Scaffold-GS-based experiments, we introduce modifications to accelerate training: the number of training iterations is set to 10k unless otherwise specified. Additionally, densification is performed only within the first 3k iterations, with a densification interval of every 20 steps after an initial 300-step warm-up. All evaluations were conducted on a single NVIDIA A100 GPU.

\subsection{Dataset and Baselines}
\label{subsec:dataset}

\textbf{Datasets.} For 4K resolution task, we evaluate our method on all nine scenes of MipNeRF-360~\cite{mip360} and two scenes from EyefulTower~\cite{eyefultower}. For details please see supplementary. We also evaluate our method in $1K$ resolution setting, where we follow the evaluation strategy of 3DGS on MipNeRF-360 and Deep Blending~\cite{hedman2018deep} datasets. For the results of Tanks and Temple~\cite{knapitsch2017tanks}, please see supplementary.

\begin{table*}[!t]
\caption{Quantitative comparison with NeRF-based methods (top half) and 3DGS-based methods in (bottom half). We compare PSNR, SSIM, and LPIPS for quality. For resource efficiency, we report training time, memory and, where applicable and peak number (Peak \#G) of Gaussians used. * denotes using Sparse Adam.  ($^{\dagger}$) denotes sensitivity-based pruning was used.}
\vspace{-5pt}
\begin{adjustbox}{width=0.98\linewidth}
\begin{tabular}{l|c|c|c|c|c|c|c|c|c|c|c|c|}
\multicolumn{1}{c}{} & \multicolumn{6}{c|}{MipNeRF-360~\cite{mip360}} & \multicolumn{6}{c}{Deep Blending~\cite{hedman2018deep}} \\ \cline{2-13} 
                             & \multicolumn{1}{c|}{SSIM$\uparrow$}           & \multicolumn{1}{c|}{PSNR$\uparrow$}           & \multicolumn{1}{c|}{LPIPS$\downarrow$}          & \multicolumn{1}{c|}{\begin{tabular}[c]{@{}c@{}}Train\\ time $\downarrow$\end{tabular}} & \multicolumn{1}{c|}{\begin{tabular}[c]{@{}c@{}}Memory $\downarrow$\end{tabular}} & \multicolumn{1}{c|}{\begin{tabular}[c]{@{}c@{}}Peak \\ \#G $\downarrow$\end{tabular}} & \multicolumn{1}{c|}{SSIM$\uparrow$}          & \multicolumn{1}{c|}{PSNR$\uparrow$}           & \multicolumn{1}{c|}{LPIPS$\downarrow$}          & \multicolumn{1}{c|}{\begin{tabular}[c]{@{}c@{}}Train\\ time $\downarrow$\end{tabular}} & \multicolumn{1}{c|}{\begin{tabular}[c]{@{}c@{}}Memory $\downarrow$\end{tabular}} & \multicolumn{1}{c|}{\begin{tabular}[c]{@{}c@{}}Peak \\ \#G $\downarrow$\end{tabular}} \\
 \hline  
Instant-NGP~\cite{instantngp} (Big)     & 0.699 & 25.59 & 0.331 & \cellcolor{Orange}7.50\,m & \cellcolor{Red}48\,MB & - & 0.817 & 24.96 & 0.390 & \cellcolor{Orange}8.0\,m & \cellcolor{Red}48\,MB & - \\
Plenoxels~\cite{plenoxels}             & 0.626 & 23.08 & 0.463 & 25.82\,m & 2.1\,GB & - & 0.795 & 23.06 & 0.510 & 27.82\,m & 2.7\,GB & - \\ \hline
Taming 3DGS~\cite{mallick2024taming}           & \cellcolor{Yellow}0.814 & \cellcolor{Yellow}27.46 & 0.218 & 15.78\,m & 628\,MB & 2.66\,M & 0.903 & 29.72 & 0.241 & 13.36\,m & 581\,MB & 2.46\,M \\
Taming 3DGS*~\cite{mallick2024taming}          & 0.809 & 26.64 & 0.227 & \cellcolor{Yellow}10.05\,m & 566\,MB & 2.39\,M & 0.902 & 29.65 & 0.248 & 8.12\,m & 543\,MB & 2.3\,M \\
Mini-Splatting~\cite{fang2024mini}        & \cellcolor{Orange}0.822 & 27.34 & 0.217 & 21.71\,m & 117\,MB & 4.23\,M & \cellcolor{Yellow}0.908 & \cellcolor{Yellow}29.90 & 0.253 & 18.05\,m & 83\,MB & 4.53\,M \\
EAGLES~\cite{Girish2023EAGLESEA}                & 0.807 & 27.09 & 0.234 & 22.22\,m & \cellcolor{Orange}57\,MB & 1.93\,M & 0.907 & 29.77 & 0.249 & 23.96\,m & \cellcolor{Orange}52\,MB & 1.96\,M \\
Mip-Splatting~\cite{Yu2023MipSplattingA3}         & \cellcolor{Red}0.828 & \cellcolor{Orange}27.64 & \cellcolor{Red}0.188 & 41.47\,m & 1\,GB & 4.18\,M & 0.903 & 29.37 & \cellcolor{Orange}0.239 & 37.26\,m & 840\,MB & 3.49\,M \\ 
Speedy-Splat~\cite{speedygs}         & 0.786 & 26.93 & 0.288 & 12.99\,m & \cellcolor{Yellow}75\,MB & 2.43\,M & 0.903 & 29.61 & 0.268 & 11.96\,m & 145\,MB & 1.77\,M \\ 

\hline
3DGS~\cite{3dgs}                  & \cellcolor{Yellow}0.814 & 27.45 & 0.217 & 30.08\,m & 640\,MB & 2.71\,M & 0.902 & 29.75 & 0.241 & 30.58\,m & 580\,MB & 2.46\,M \\
Turbo-3DGS (Ours)       & \cellcolor{Yellow}0.814 &27.39 & \cellcolor{Yellow}0.205 & 12.18m &861\,MB & 3.64~M &0.905 & 29.21& \cellcolor{Yellow}0.240 & 12.13\,m & 667\,MB& 2.82~M\\ 
   Turbo-3DGS$^{\dagger}$ (Ours)    & 0.798 &27.06 & 0.253 & 12.10\,m &287\,MB & \cellcolor{Yellow}1.29~M & 0.907 & 29.76 & 0.260 & \cellcolor{Yellow}8.03\,m & 219\,MB& \cellcolor{Yellow}0.95~M\\ \hline
Scaffold-GS~\cite{scaffoldgs}           & \cellcolor{Yellow}0.814 & \cellcolor{Red}27.71 & 0.221 & 23.85\,m & 180\,MB & \cellcolor{Red}0.59\,M & \cellcolor{Orange}0.909 & \cellcolor{Orange}30.29 & 0.252 & 17.25\,m & \cellcolor{Yellow}54\,MB & \cellcolor{Red}0.18\,M \\
Turbo-Scaffold-GS (Ours)       &  \cellcolor{Yellow}0.814 & 27.43 & \cellcolor{Orange}0.208 & \cellcolor{Red}6.86\,m & 203\,MB& \cellcolor{Orange}0.67\,M & \cellcolor{Red}0.911 & \cellcolor{Red}30.48 & \cellcolor{Red}0.233 & \cellcolor{Red}4.0\,m & 156\,MB & \cellcolor{Orange}0.51\,M \\
\hline
\end{tabular}
\end{adjustbox}
\vspace{-10pt}
\label{tab:quant_table}
\end{table*}

\vspace{5pt}
\noindent
\textbf{Metrics.} We use standard quality metrics Peak Signal-to-Noise Ratio (PSNR), Structural Similarity (SSIM)~\cite{wang2004image} and Learned Perceptual Image Patch Similarity (LPIPS)~\cite{zhang2018unreasonable} to measure the visual quality. In the context of Gaussian Splatting, we recommend prioritizing LPIPS, as it most closely aligns with human perceptual experience. Further, we also report the optimization time, memory consumption and peak number of Gaussian primitives ($\#G$) during training.

\subsection{Results and Analysis}
\label{subsec:results}

\begin{figure*}[t]
  \centering
  \makebox[\textwidth][c]{
    \begin{tabular}{@{\hskip 0pt}c@{\hskip 2pt}c@{\hskip 2pt}c@{\hskip 2pt}c@{\hskip 2pt}c@{\hskip 0pt}}
        Ground Truth & 3DGS & Scaffold-GS & Taming 3DGS & Turbo-GS (Ours) \\
         \includegraphics[width=0.2\textwidth]{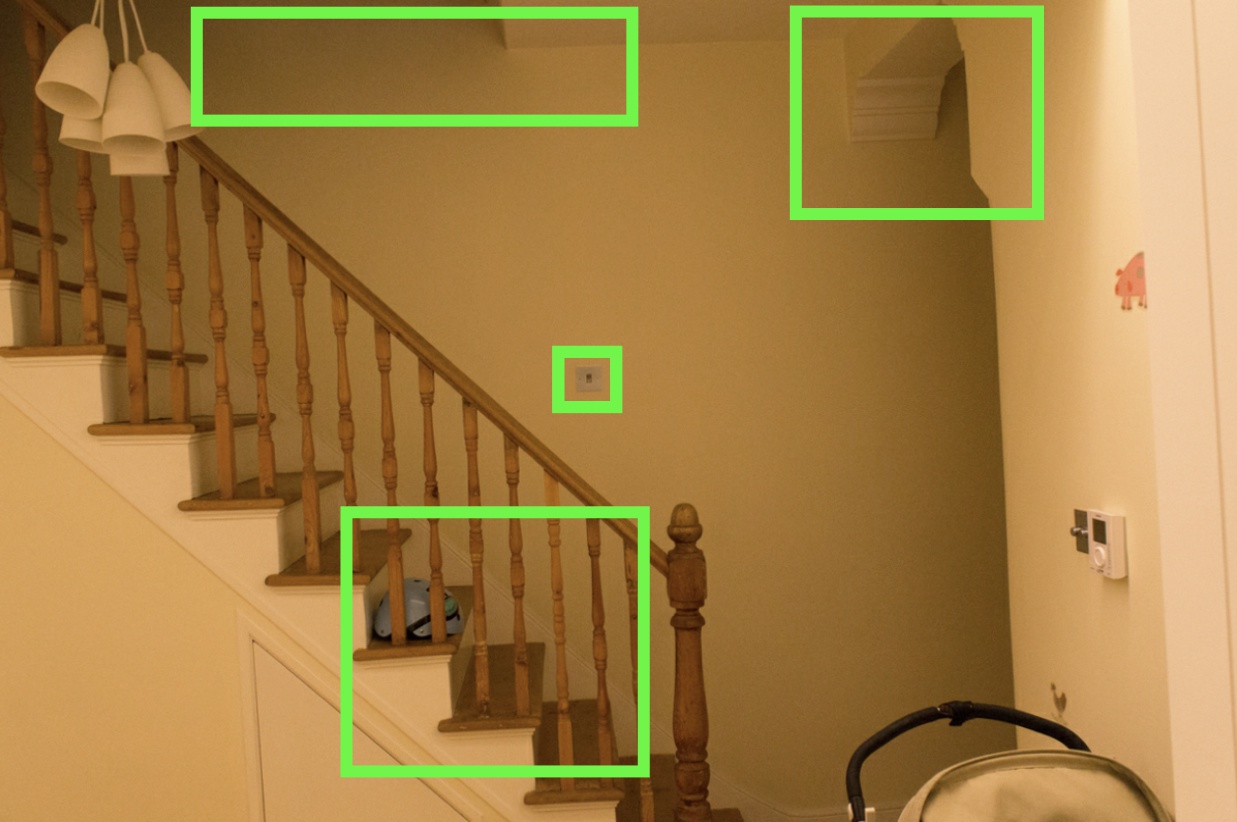} & 
         \includegraphics[width=0.2\textwidth]{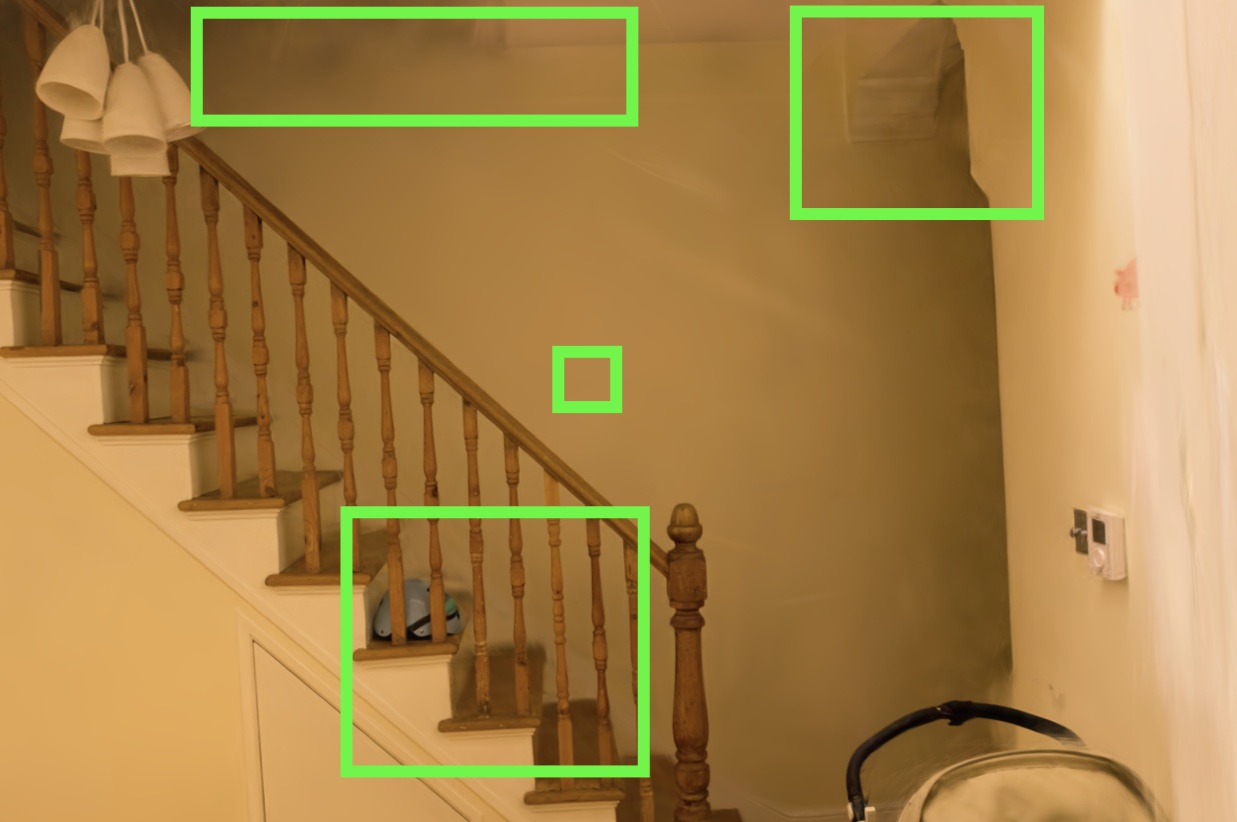} &
         \includegraphics[width=0.2\textwidth]{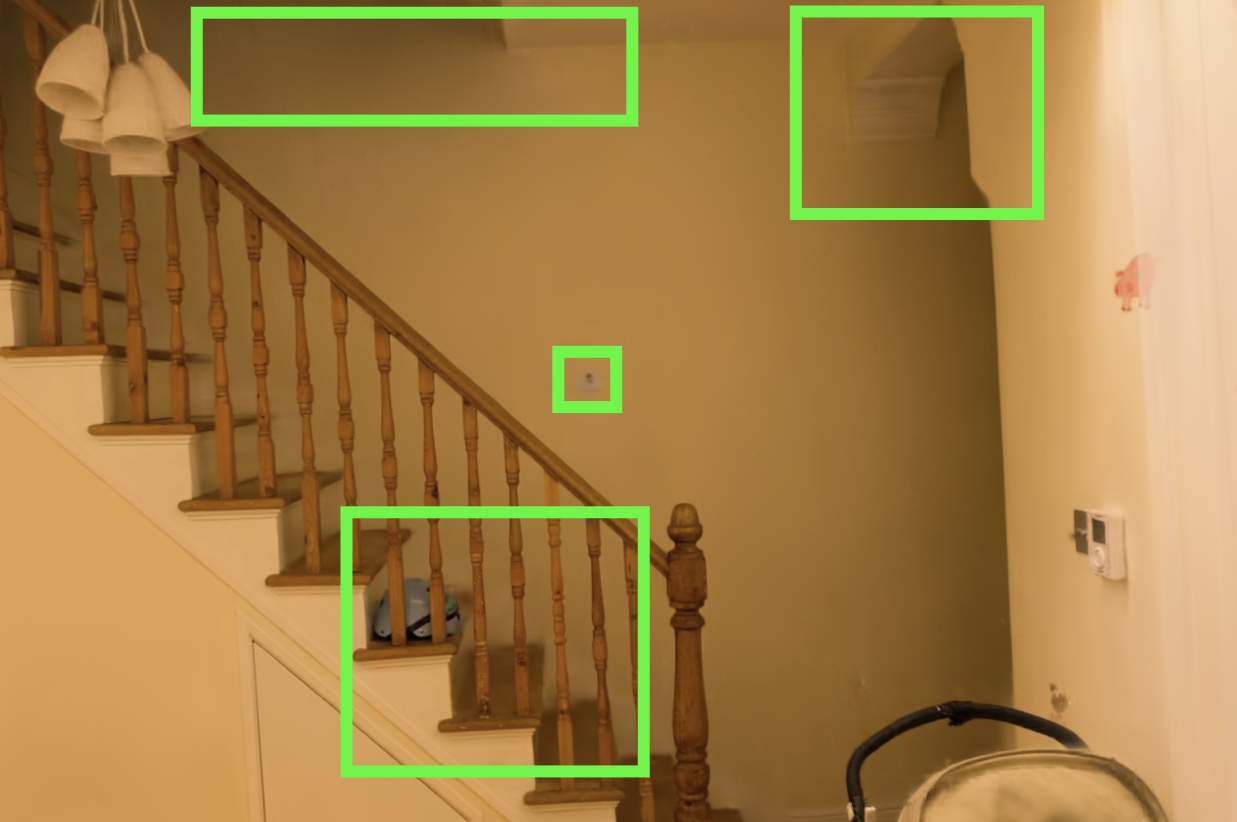} & 
         \includegraphics[width=0.2\textwidth]{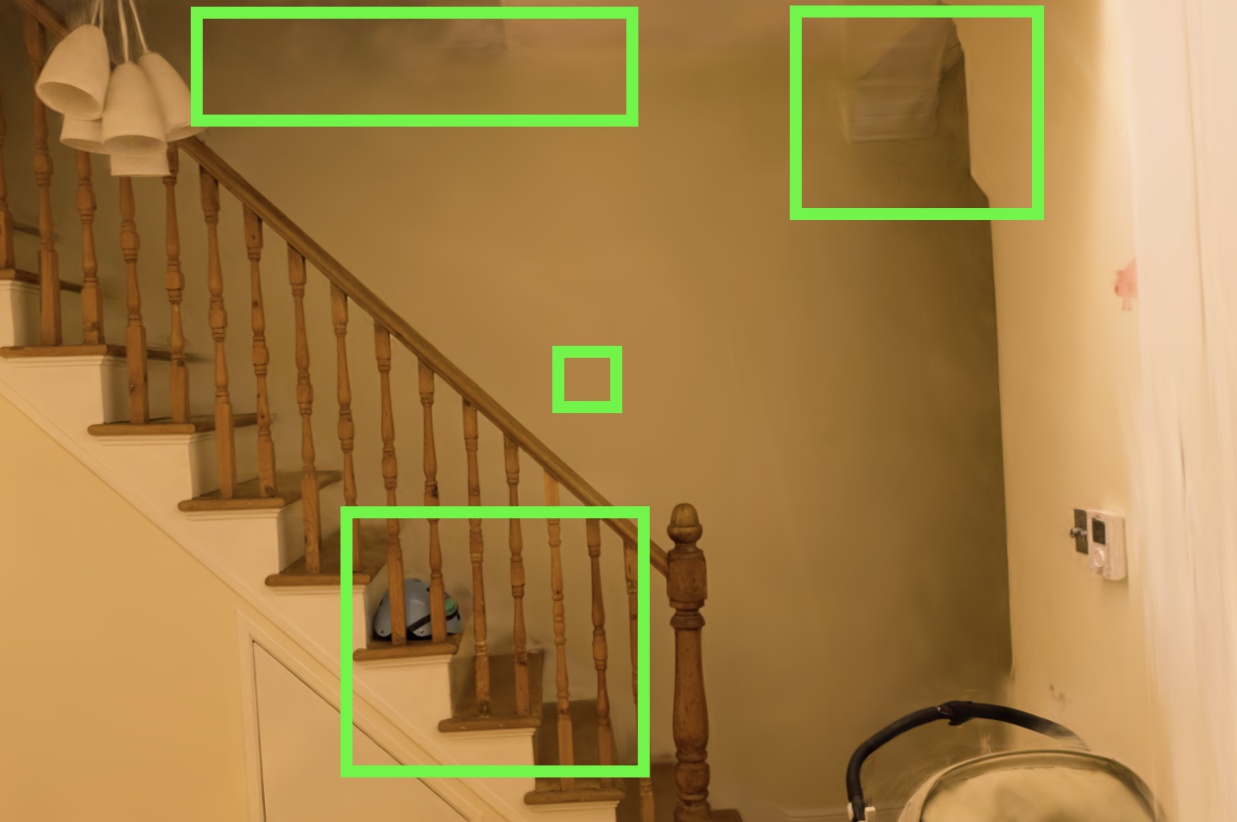} & 
         \includegraphics[width=0.2\textwidth]{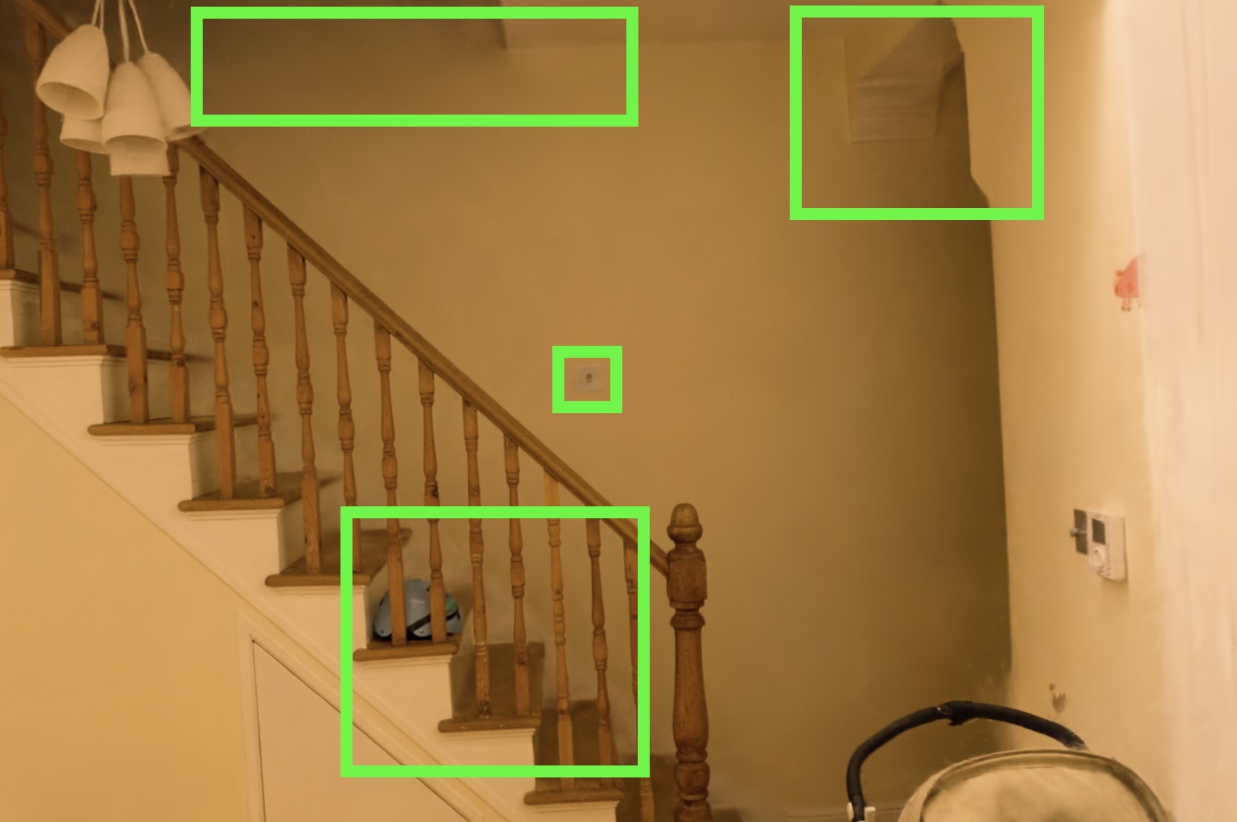} \\
         \includegraphics[width=0.2\textwidth]{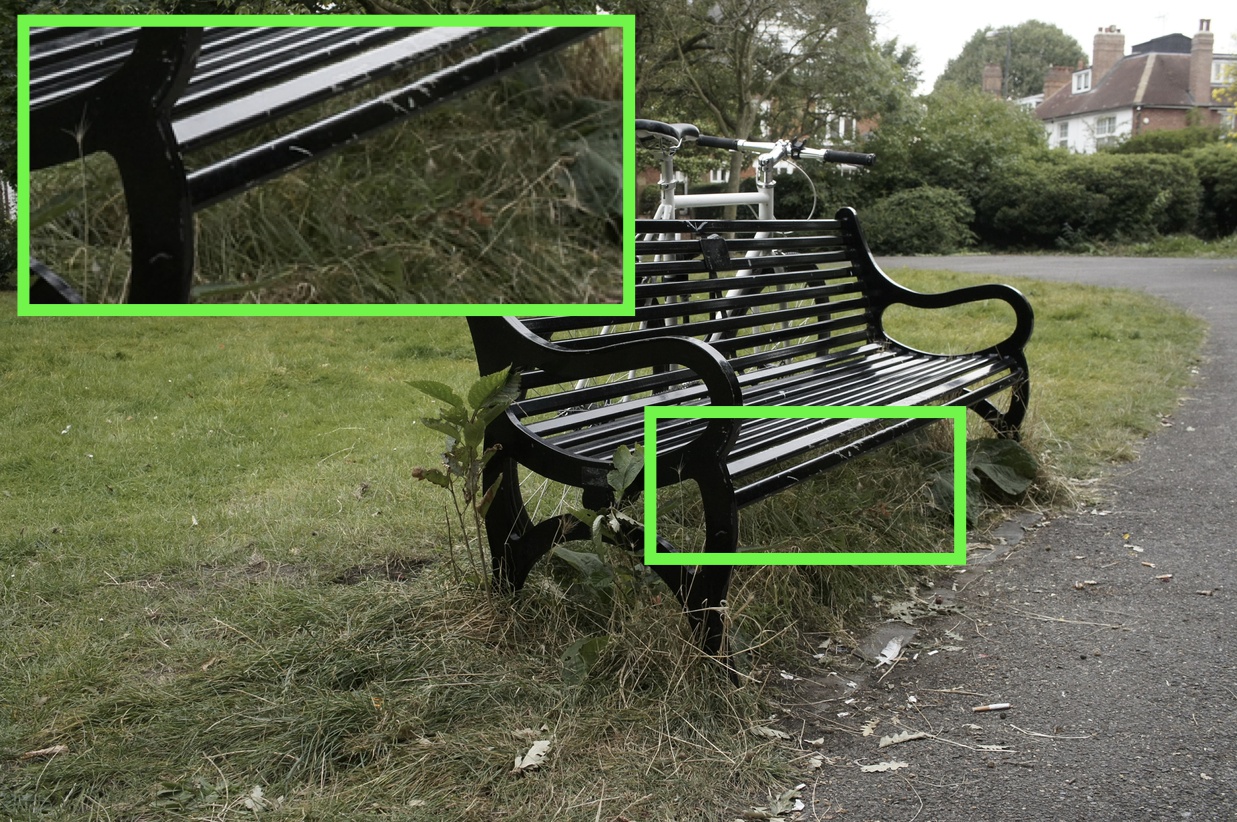} & 
         \includegraphics[width=0.2\textwidth]{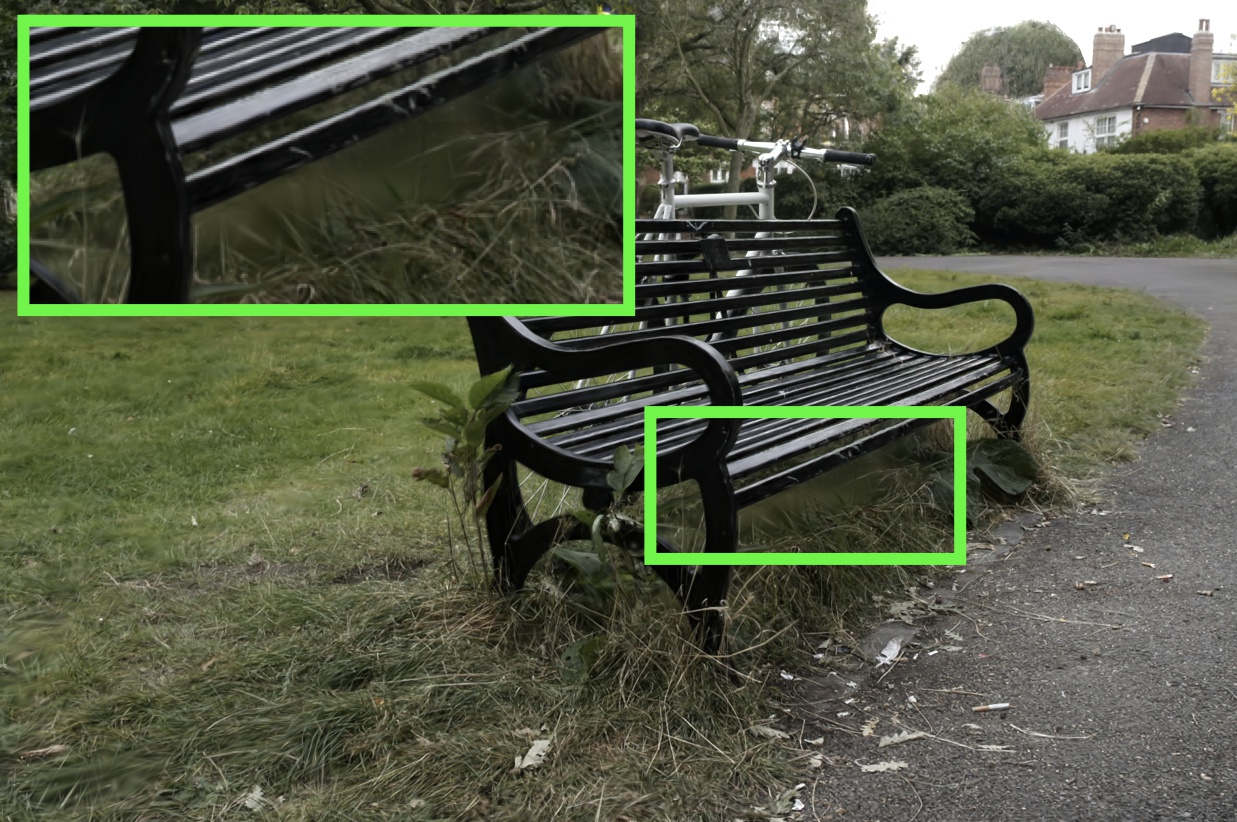} &
         \includegraphics[width=0.2\textwidth]{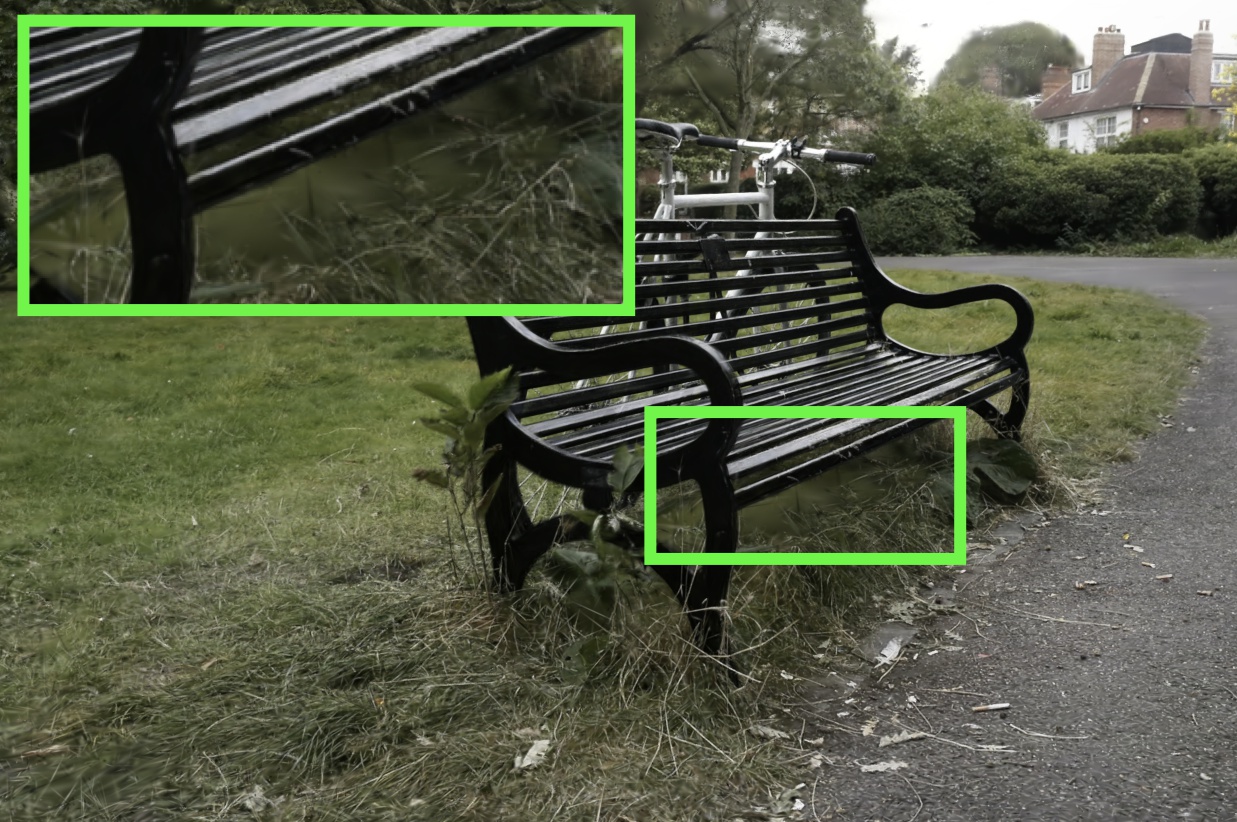} & 
         \includegraphics[width=0.2\textwidth]{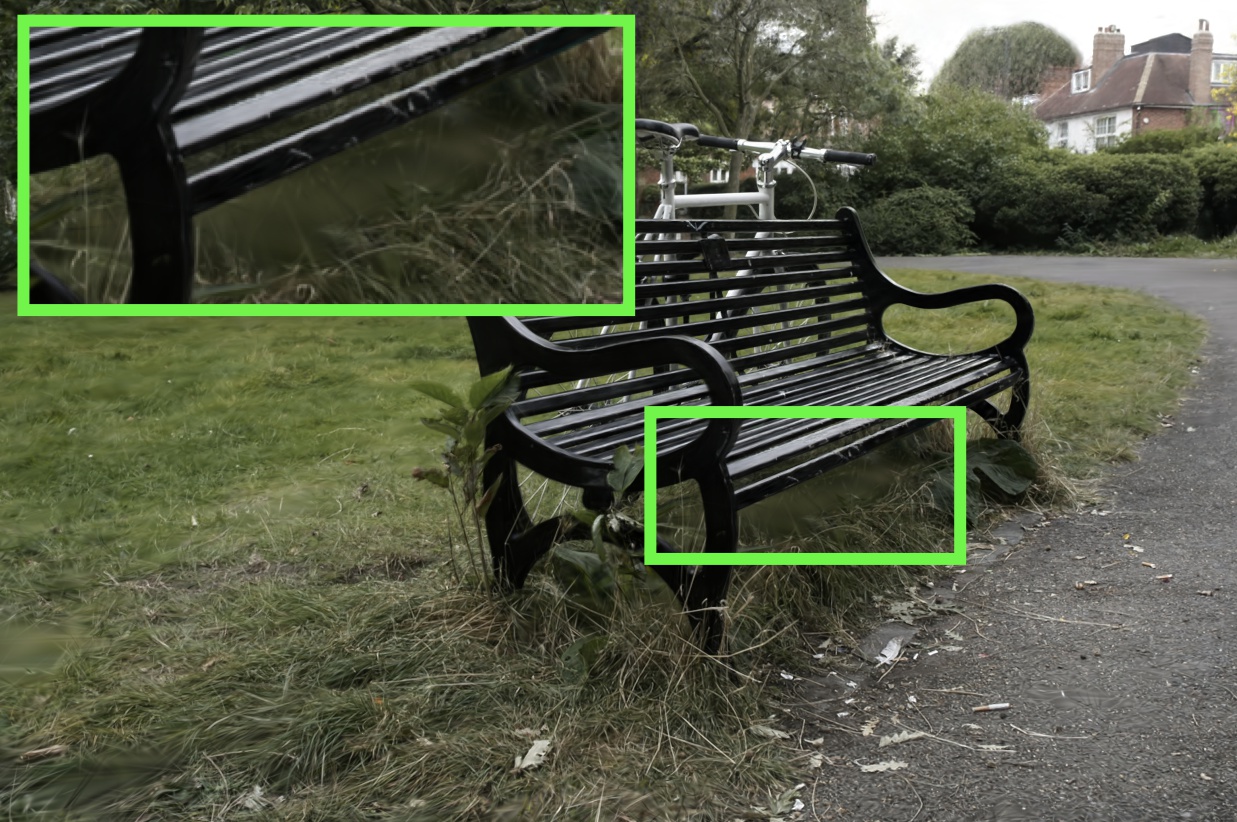} & 
         \includegraphics[width=0.2\textwidth]{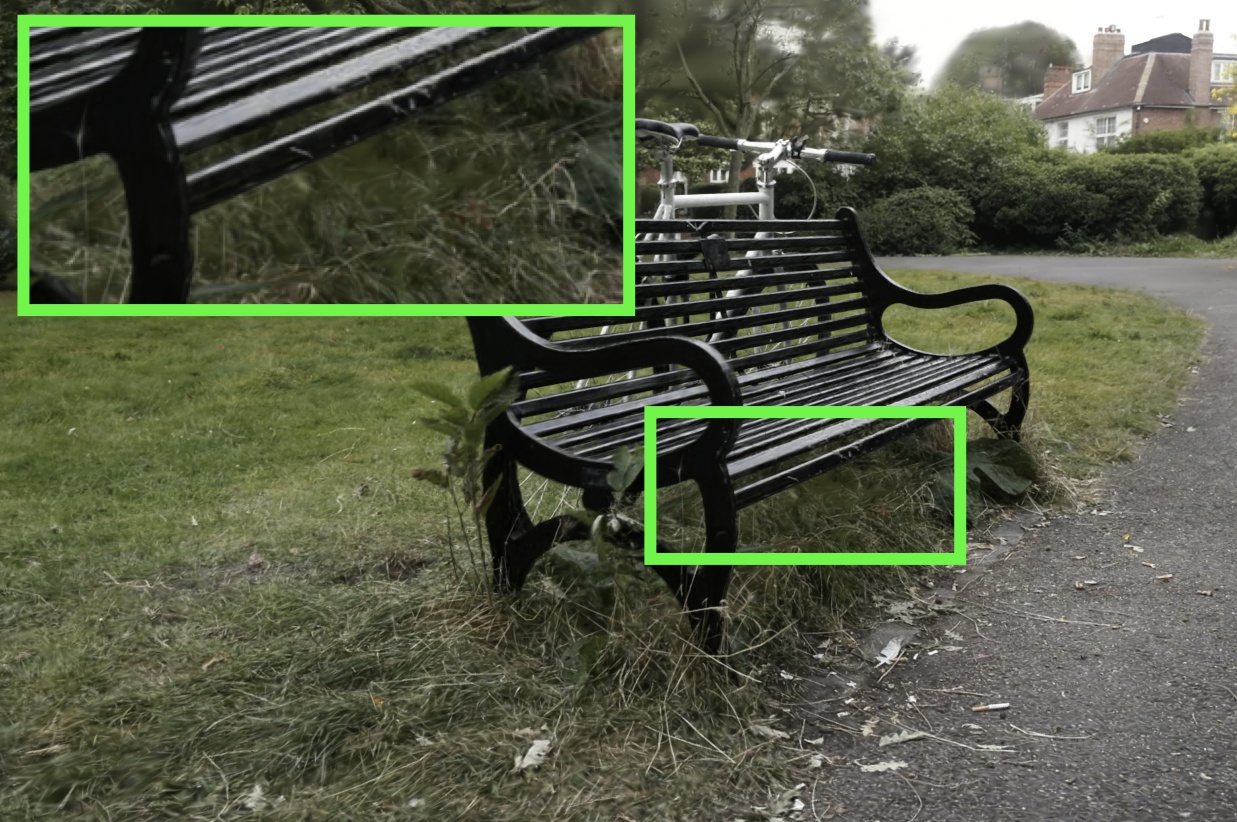} \\
         \includegraphics[width=0.2\textwidth]{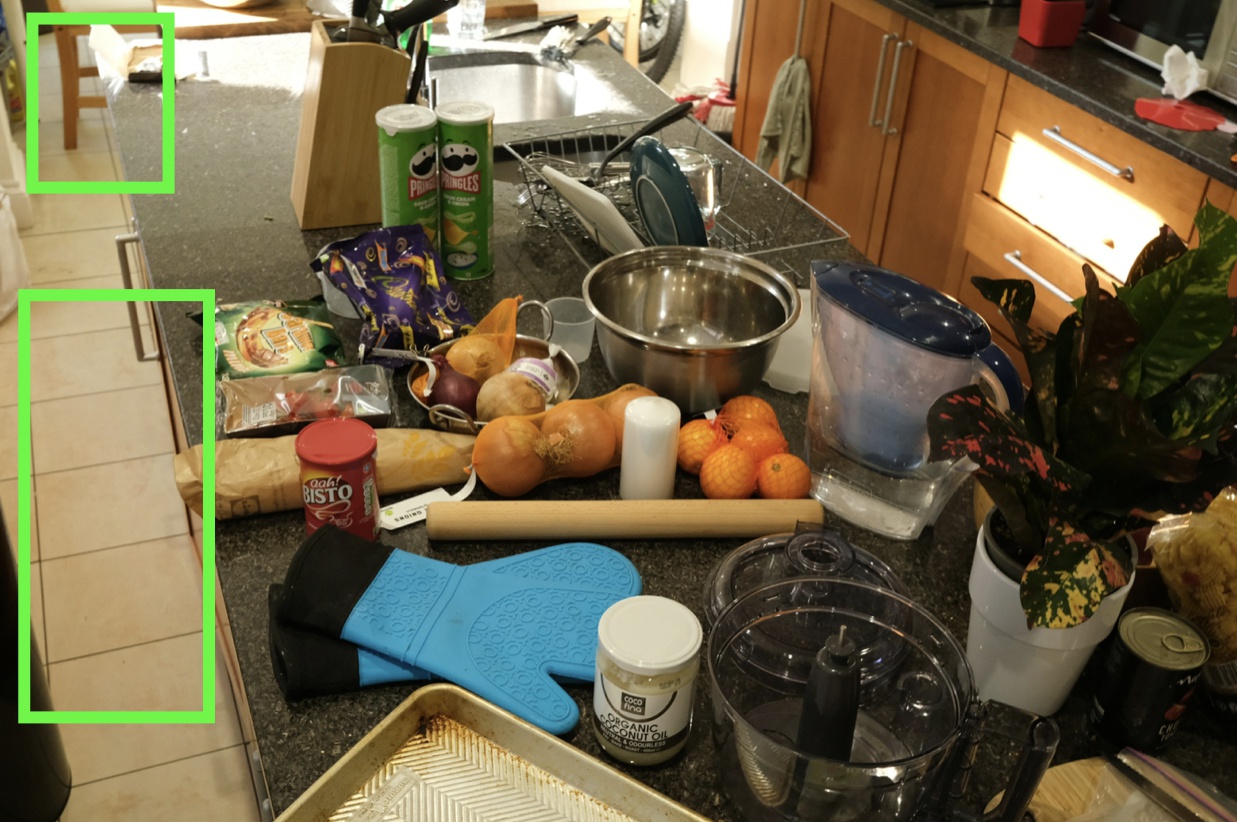} & 
         \includegraphics[width=0.2\textwidth]{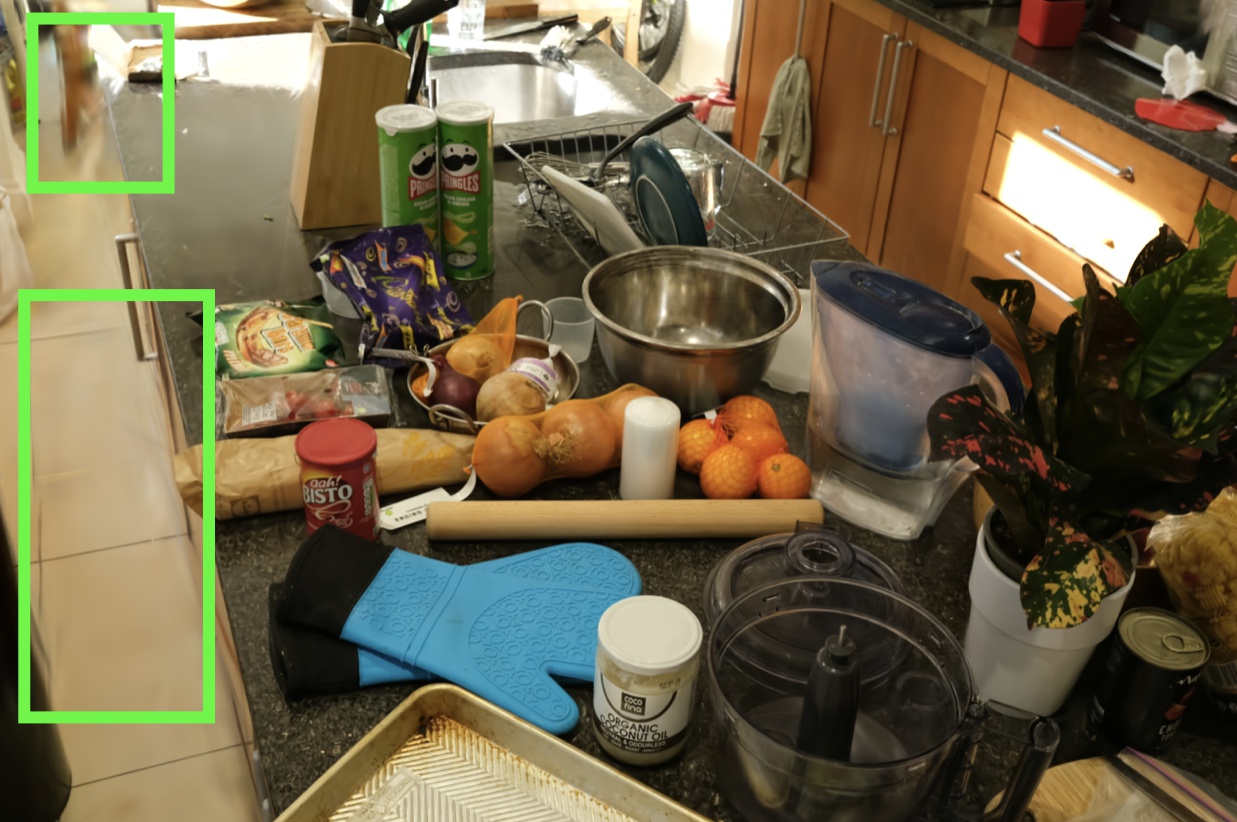} &
         \includegraphics[width=0.2\textwidth]{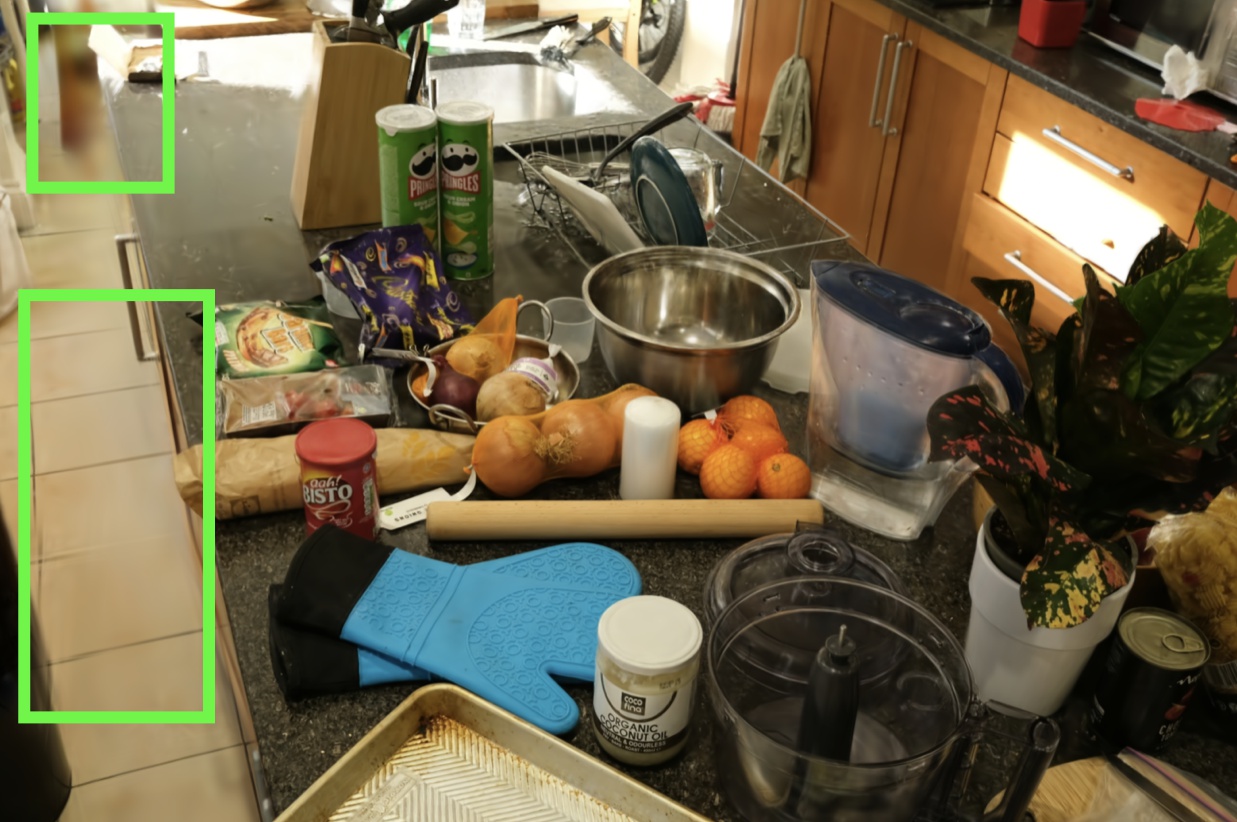} & 
         \includegraphics[width=0.2\textwidth]{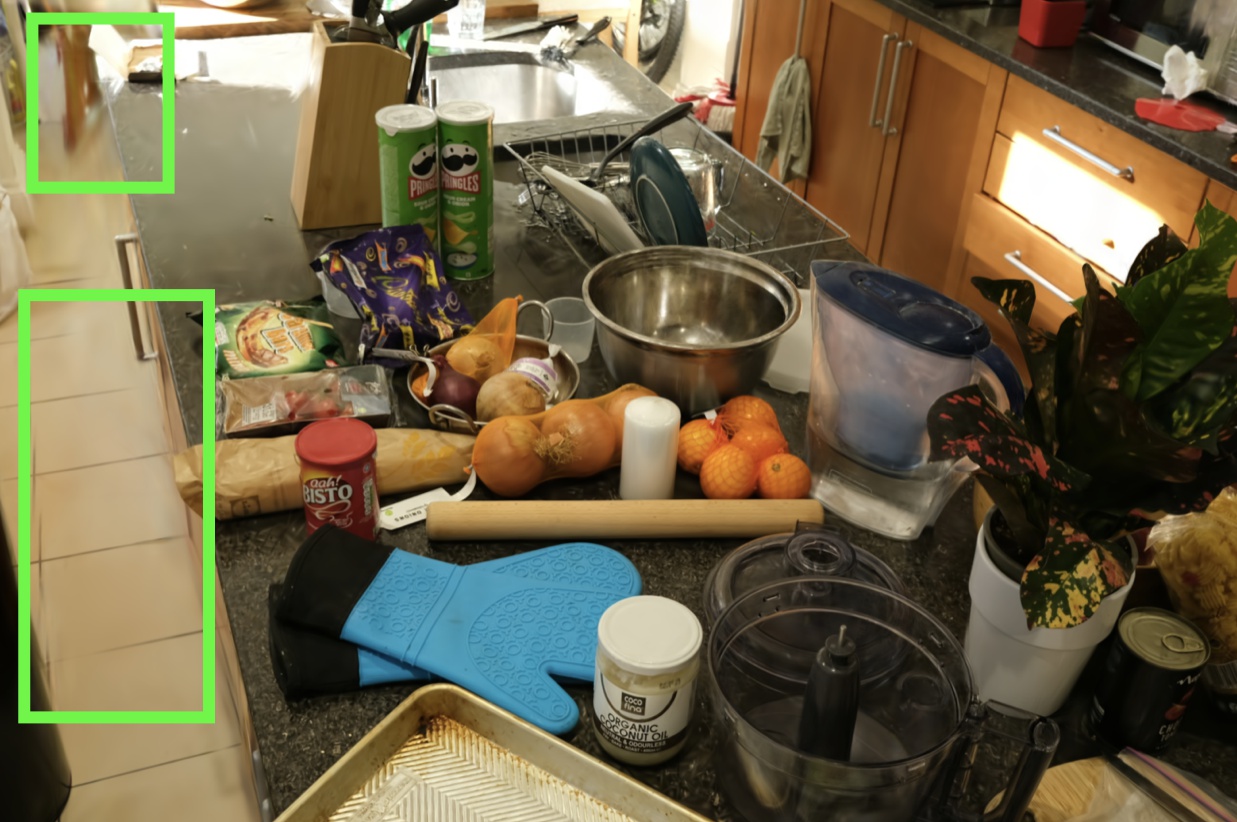} & 
         \includegraphics[width=0.2\textwidth]{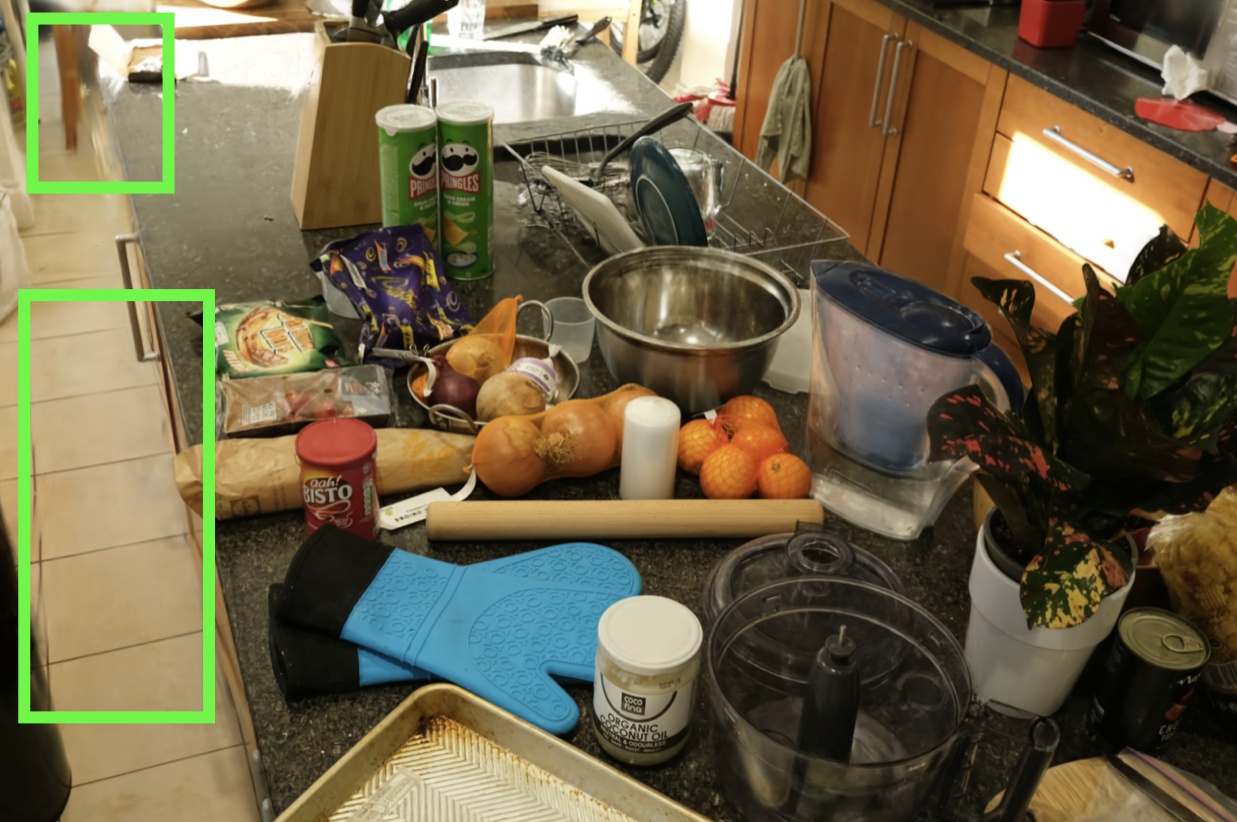} \\
         \includegraphics[width=0.2\textwidth]{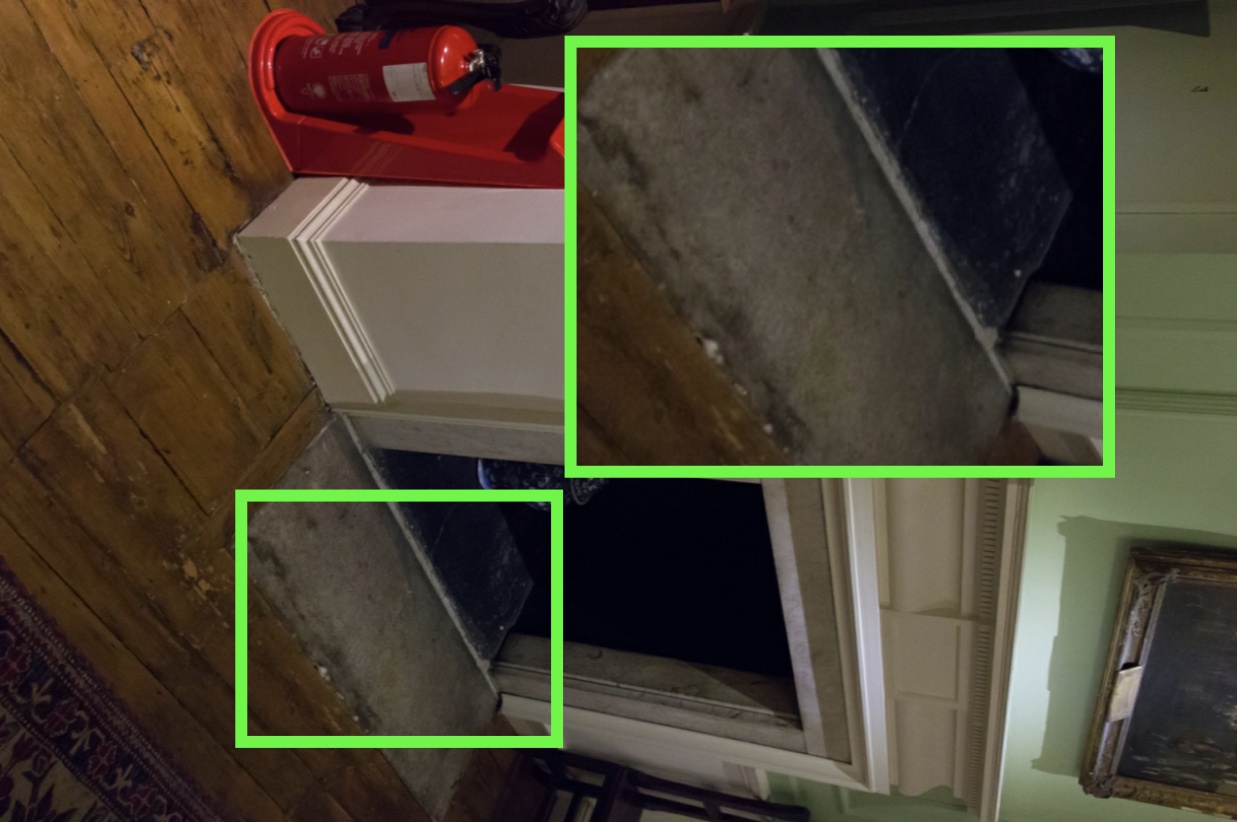} & 
         \includegraphics[width=0.2\textwidth]{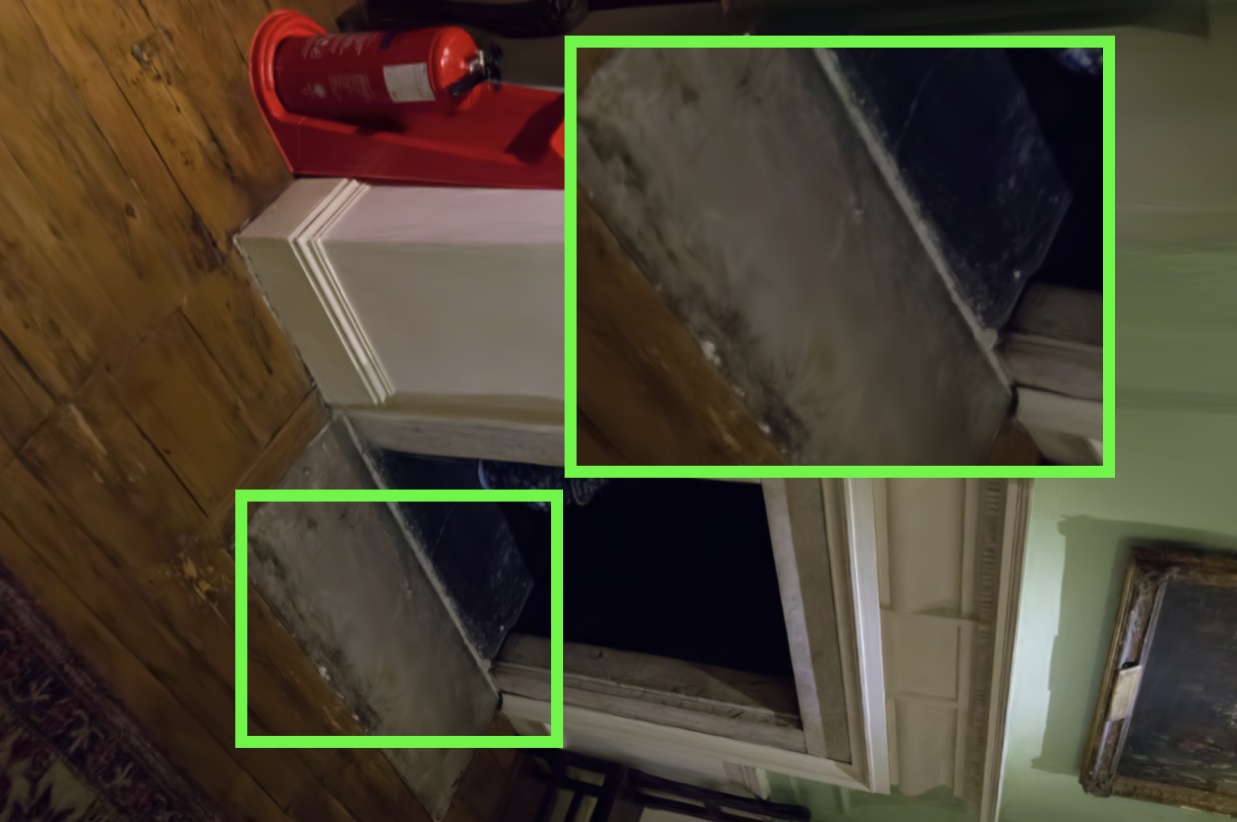} &
         \includegraphics[width=0.2\textwidth]{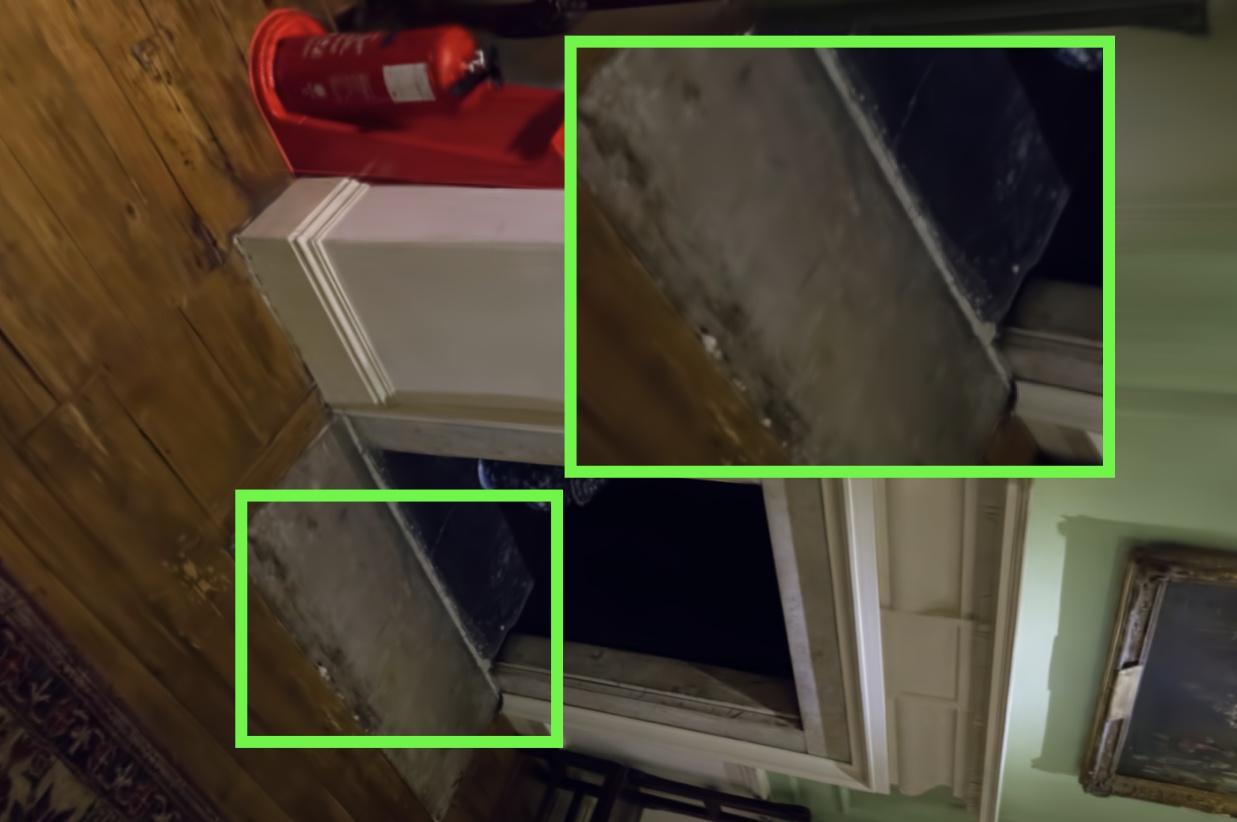} & 
         \includegraphics[width=0.2\textwidth]{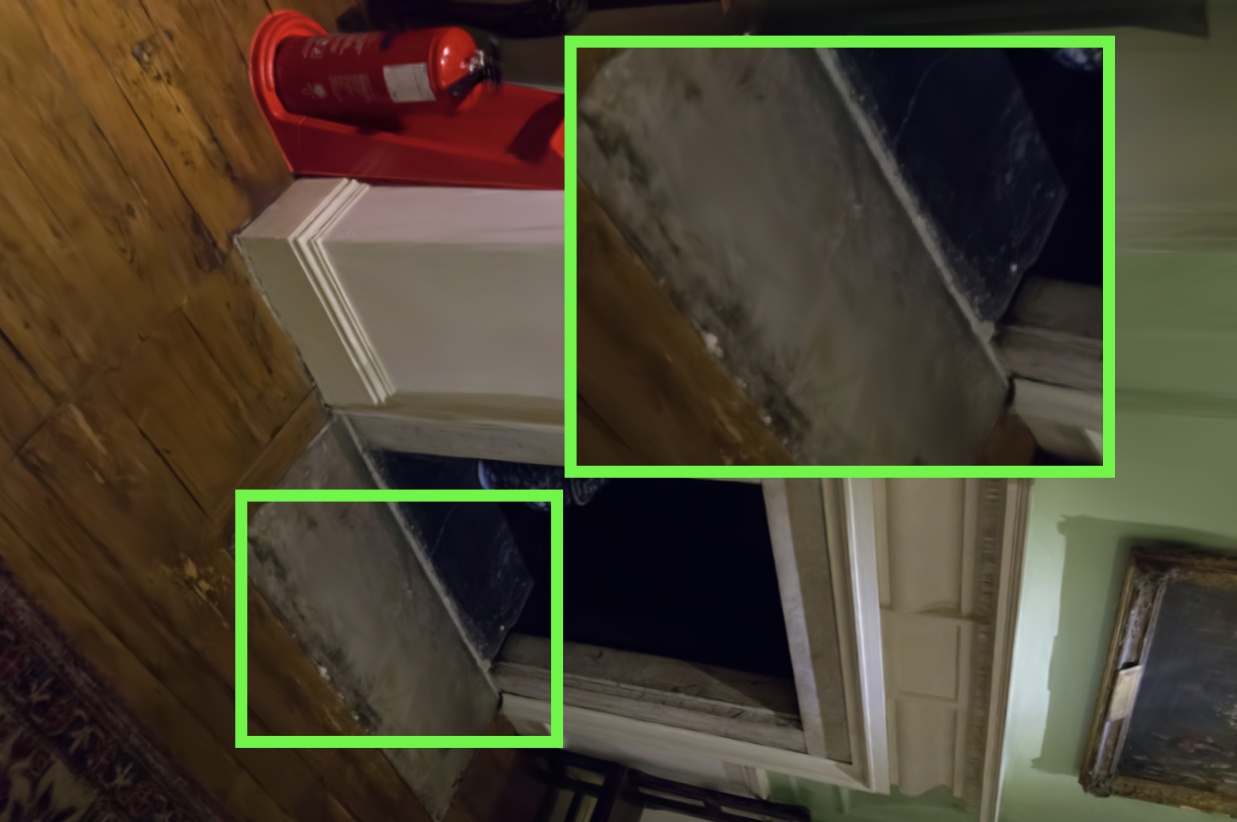} & 
         \includegraphics[width=0.2\textwidth]{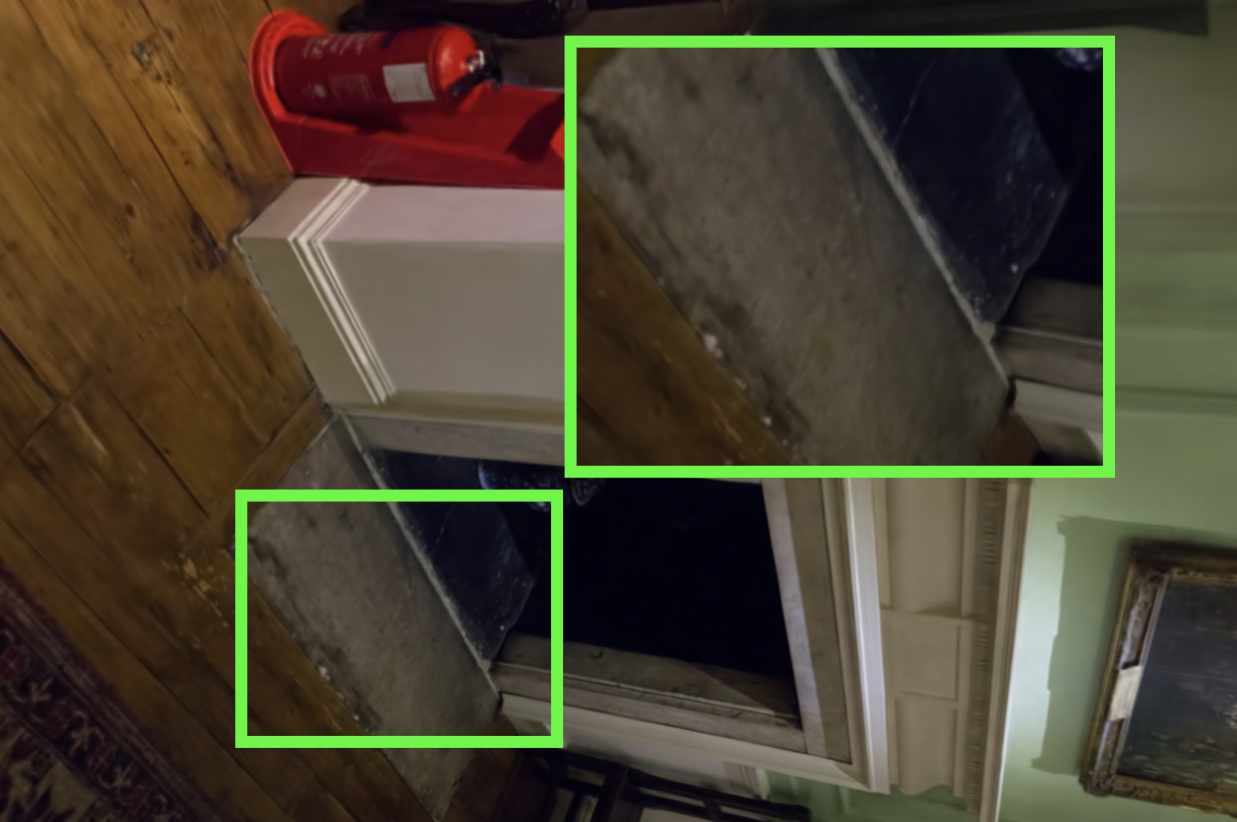} \\
         \includegraphics[width=0.2\textwidth]{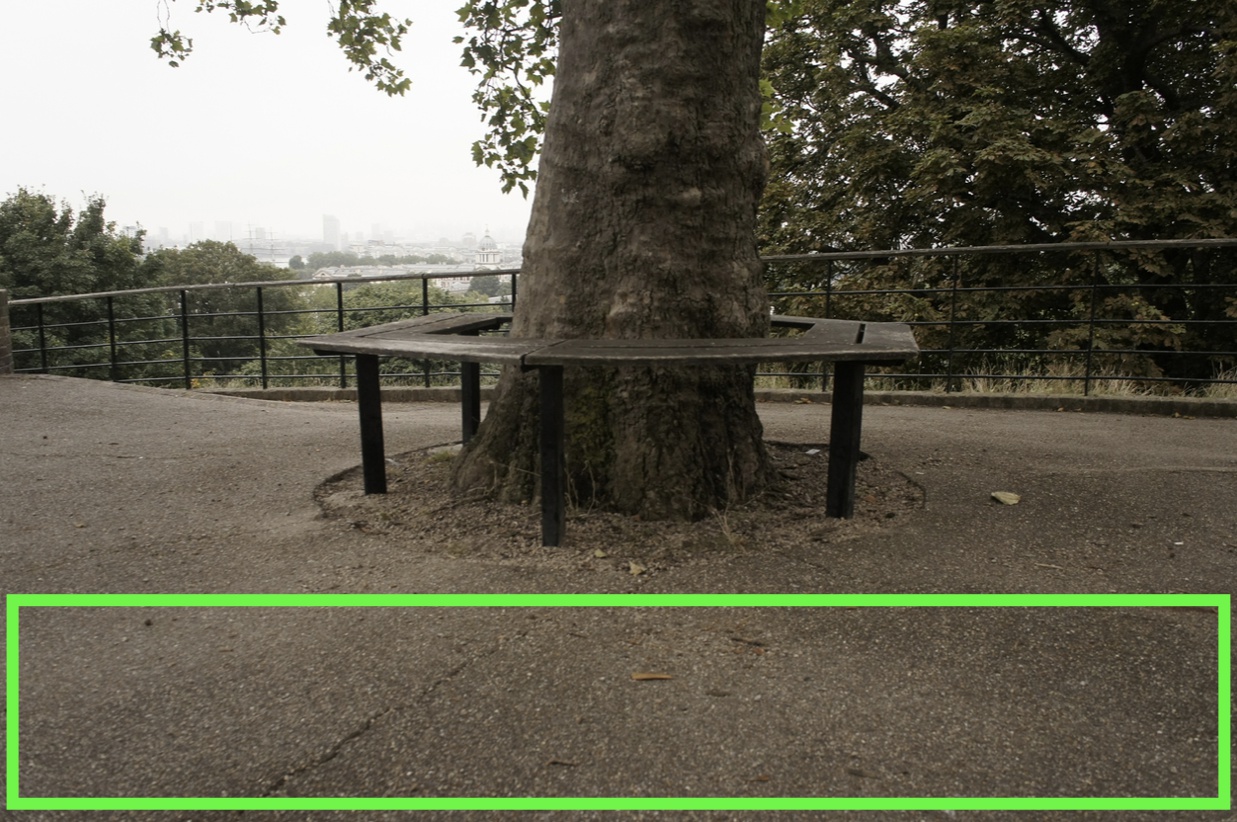} & 
         \includegraphics[width=0.2\textwidth]{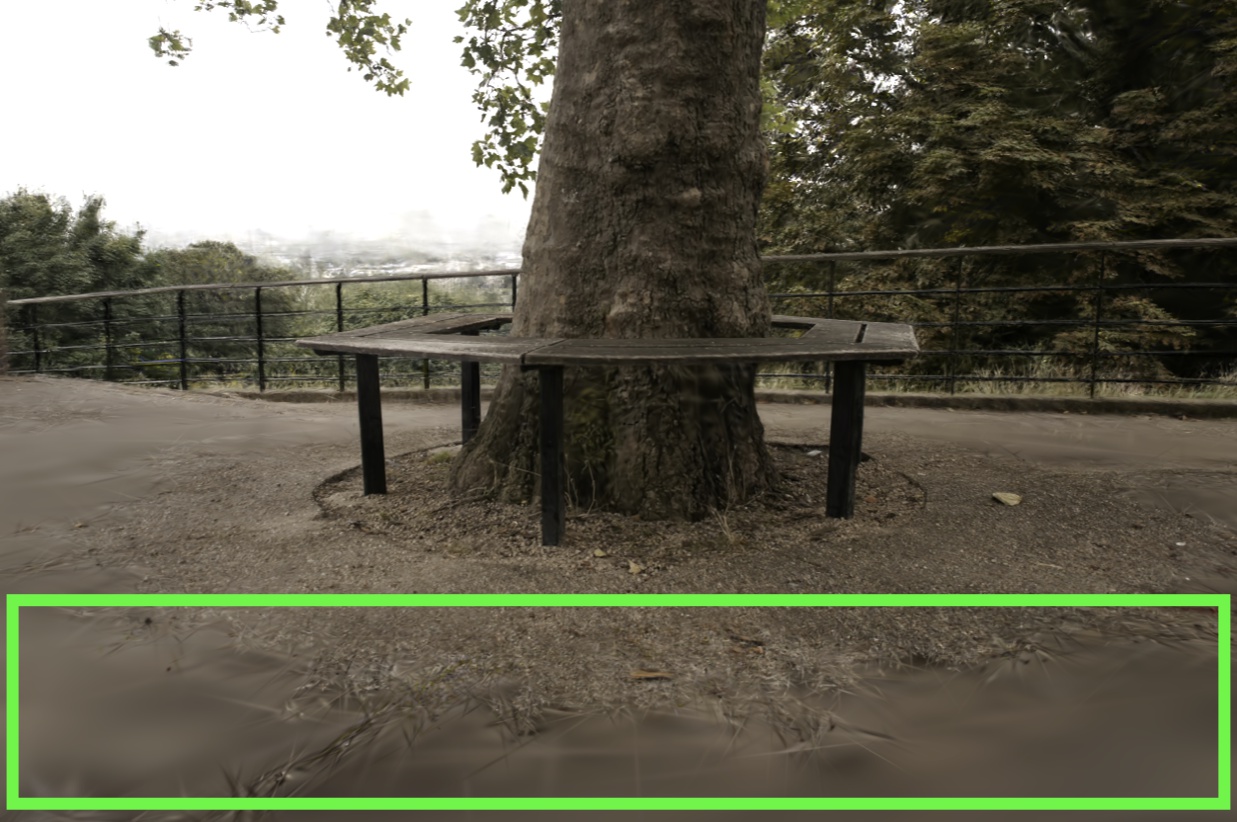} &
         \includegraphics[width=0.2\textwidth]{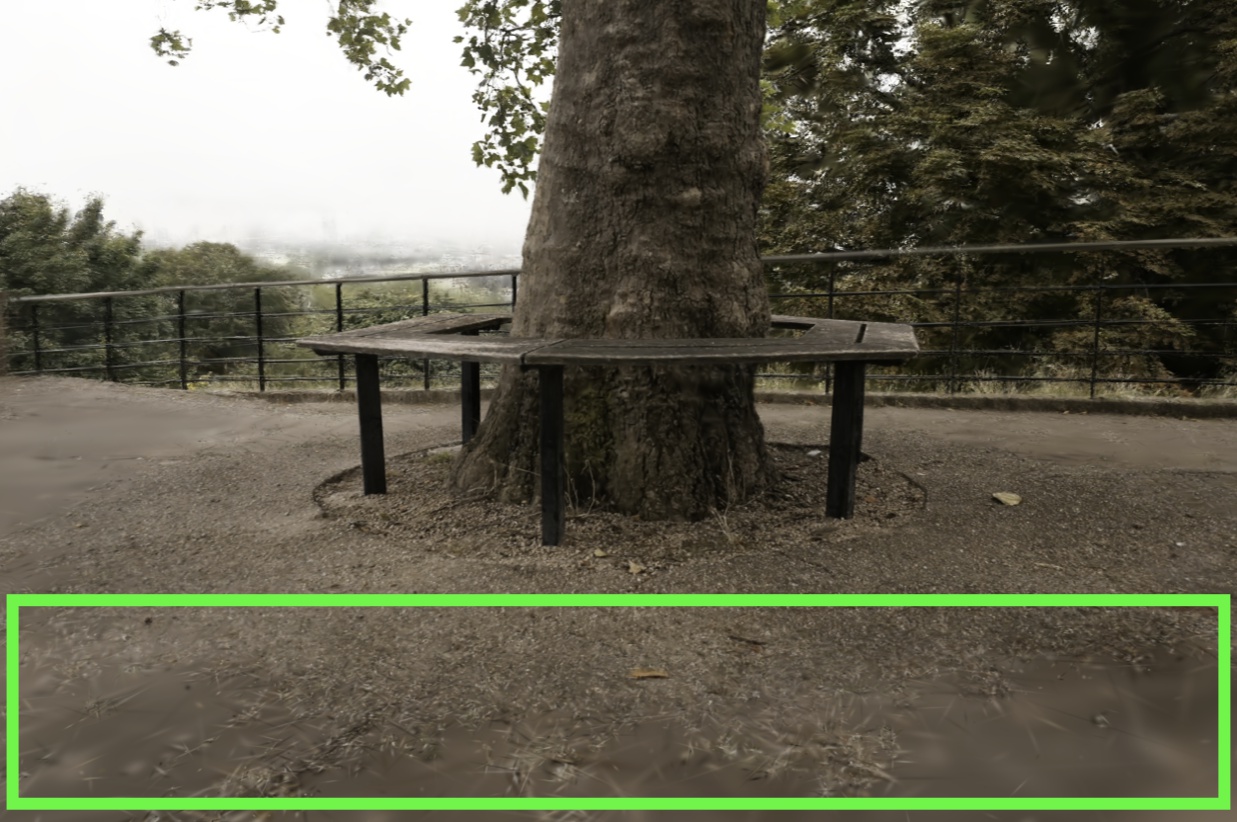} & 
         \includegraphics[width=0.2\textwidth]{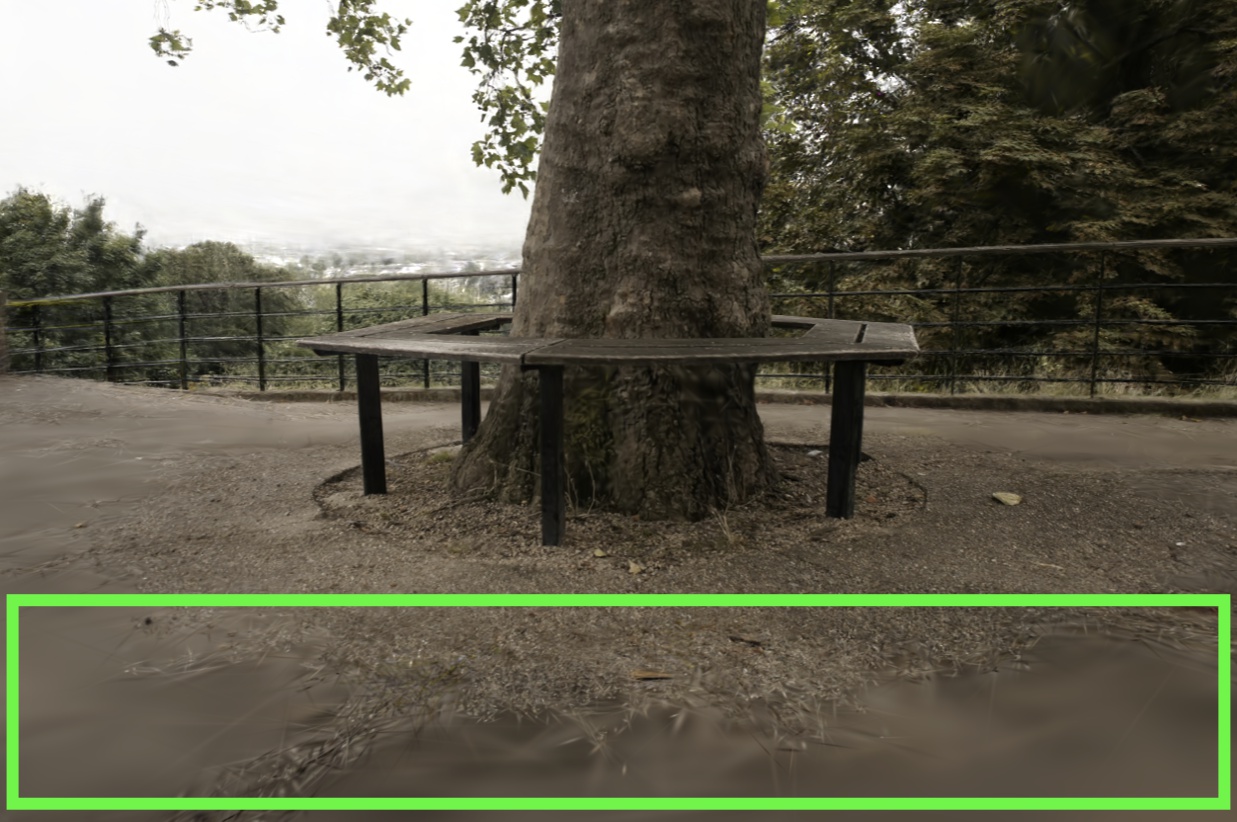} & 
         \includegraphics[width=0.2\textwidth]{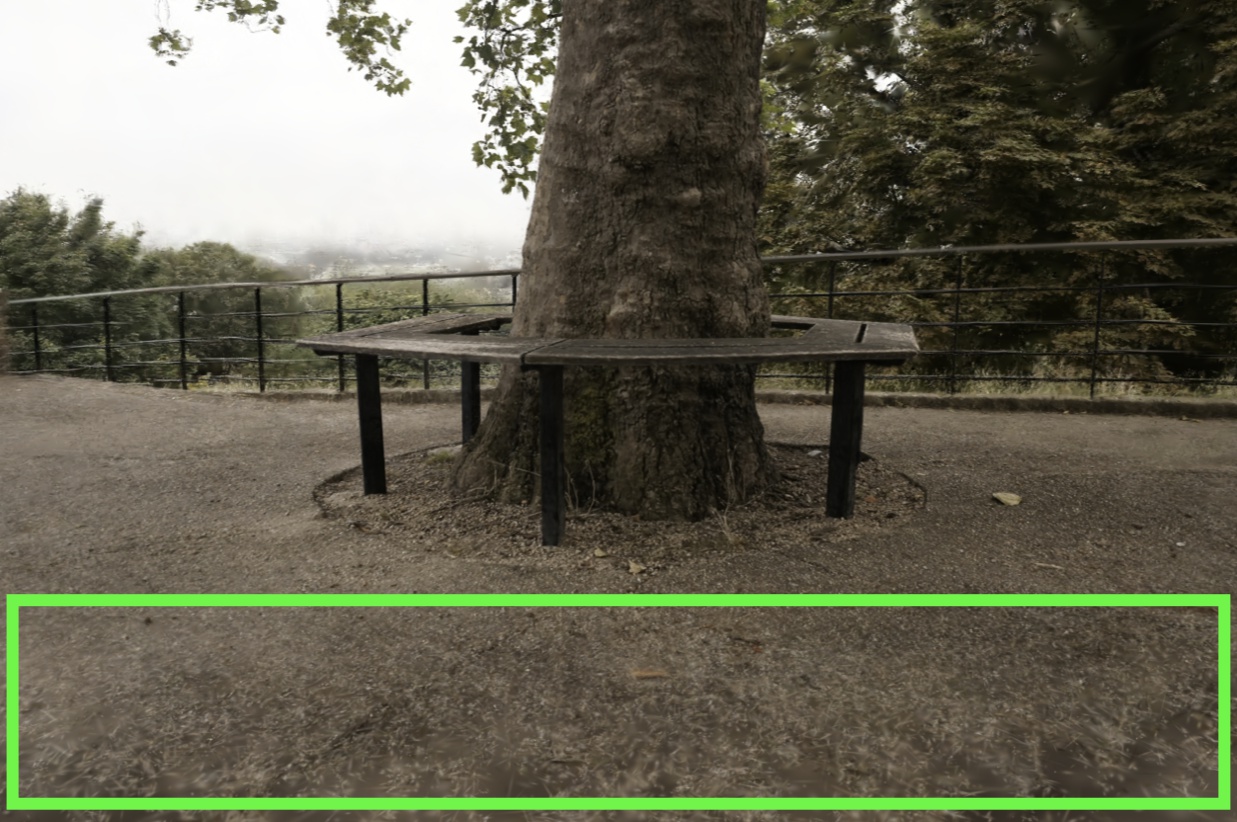} \\
         \includegraphics[width=0.2\textwidth]{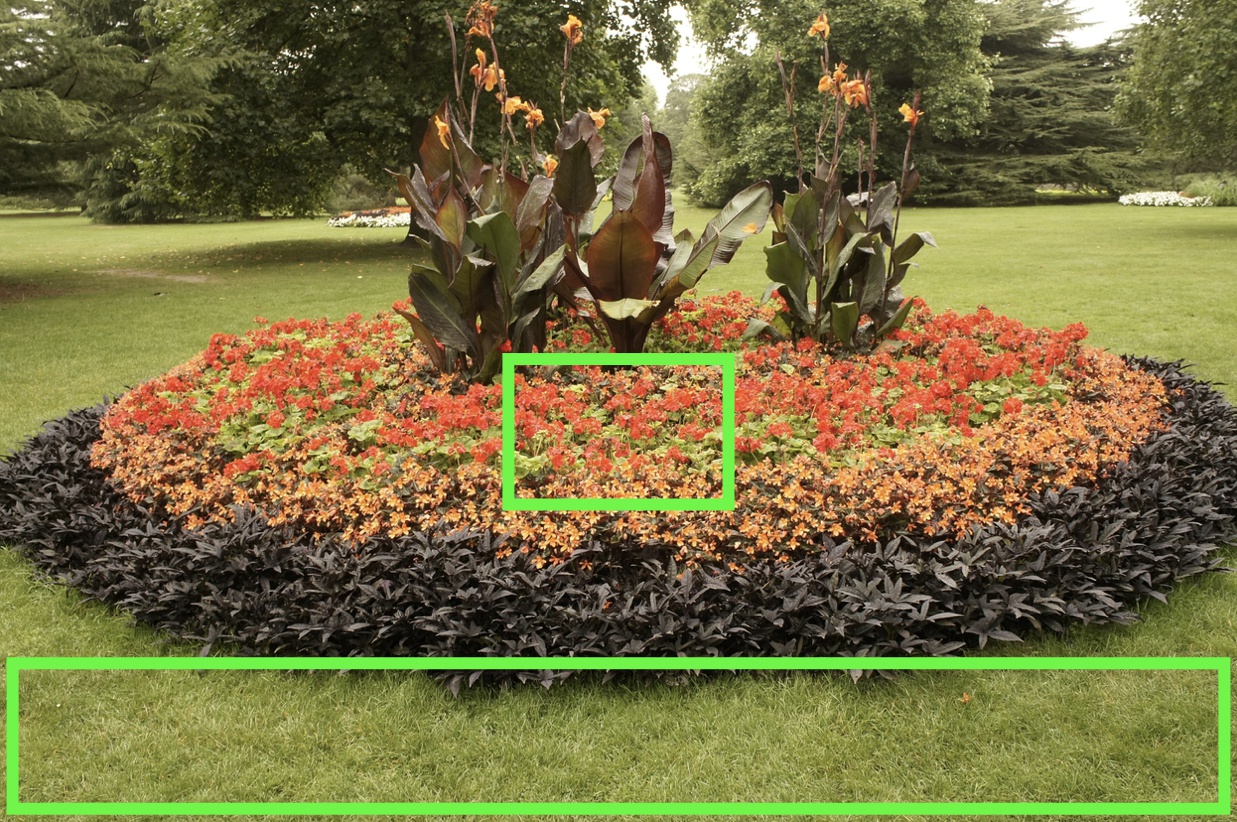} & 
         \includegraphics[width=0.2\textwidth]{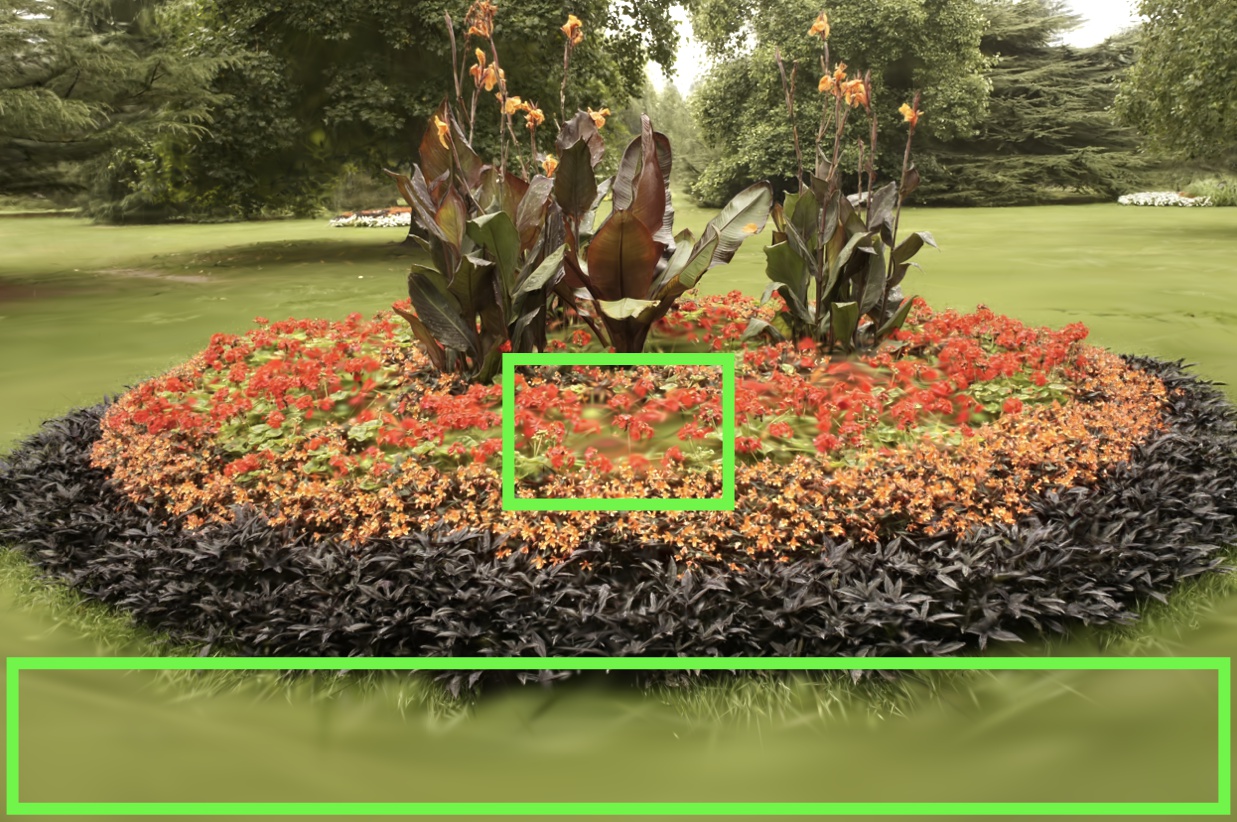} &
         \includegraphics[width=0.2\textwidth]{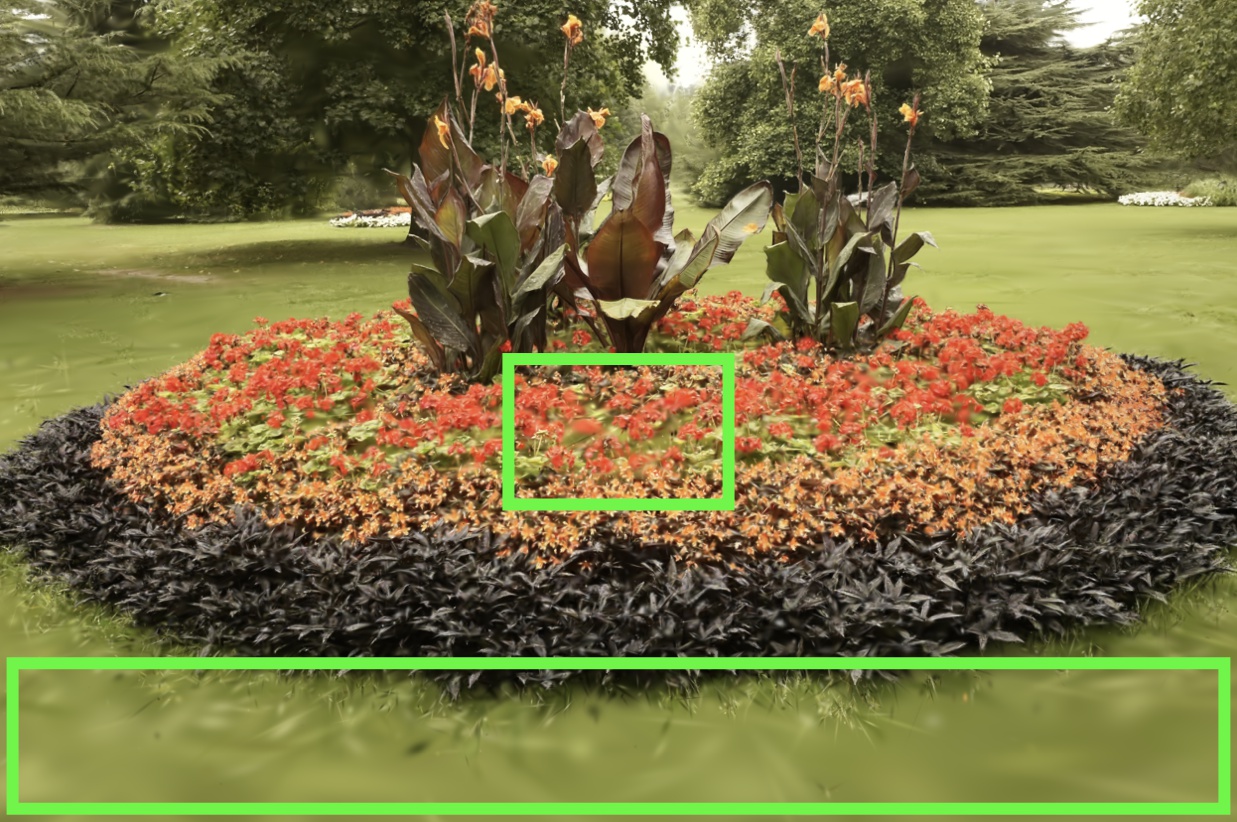} & 
         \includegraphics[width=0.2\textwidth]{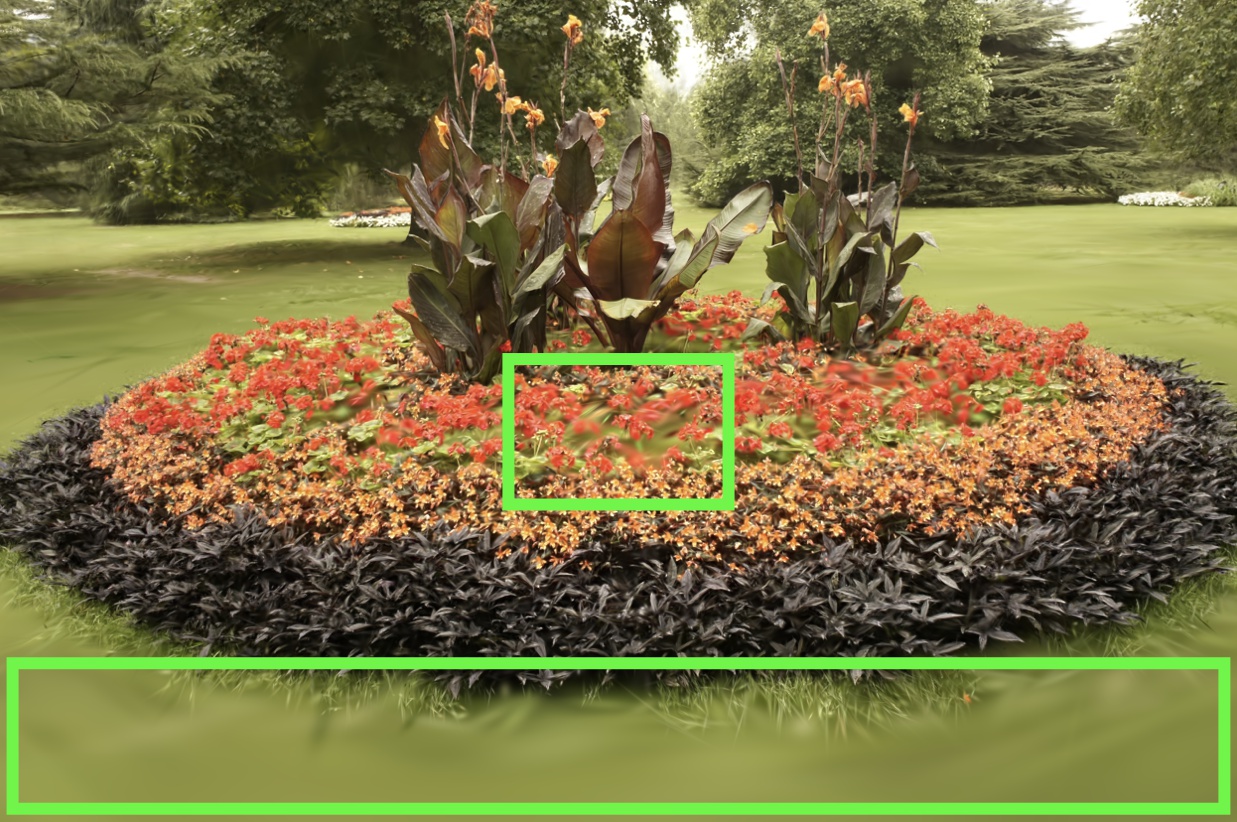} & 
         \includegraphics[width=0.2\textwidth]{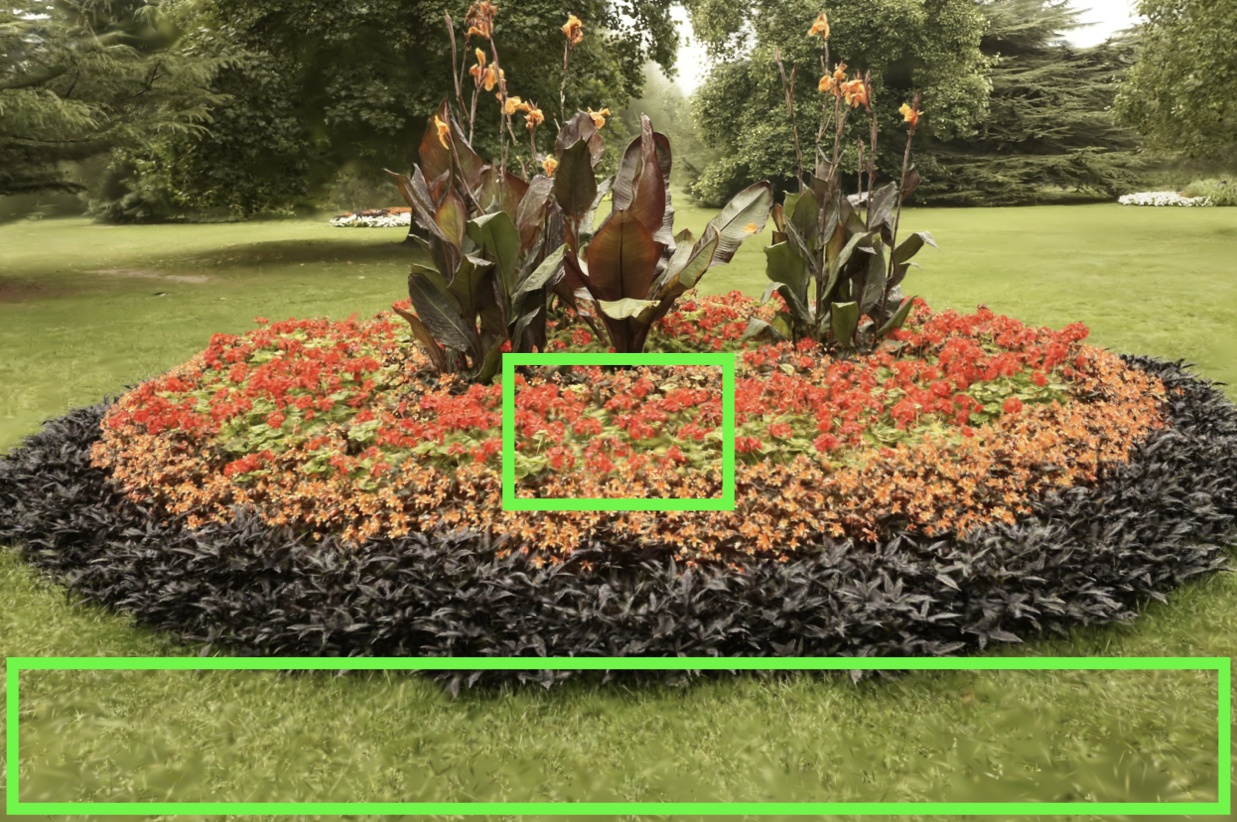} \\
    \end{tabular}
  }
  \vspace{-5pt}
  \caption{Qualitative comparison with prior 3DGS-based methods and the corresponding ground truth images from testing views. We obtain on-par or better results compared to 3DGS, Scaffold-GS, and Taming 3DGS, especially in high-frequency and texture-less areas highlighted in green: grass (second, last row), floor (fourth row) and pavement (fifth row).}
  \vspace{-10pt}
  \label{fig:qualitative_results}
\end{figure*}

\paragraph{Quantitative Analysis.}
On 4K dataset from MipNeRF-360~\cite{mip360} and EyefulTower~\cite{eyefultower} dataset, in~\cref{tab:quant_table_4k} we show the comparison results with original baselines, the enhanced baselines which have already combined with most powerful updates in this community. Results show that we could finish training a 4K scene using around 10 minutes. This represents a significant breakthrough, as in previous training a 1K scene would take around half an hour. Our method maintains consistently faster training speed in 4K data than the baseline due to the dilated rendering, implying that we find another dimension for optimization the Gaussian training process, e.g., resolution. And, we achieve a higher rendering quality, especially in the LPIPS metric, a more important measurement for gaussian-based visual quality. The improvement in visual quality comes from the optimization in densification strategy, as we will prove in the ablation. We also provide comparisons on the most popular low-resolution settings, as shown in~\cref{tab:quant_table}. Following evaluation methodology proposed in 3DGS~\cite{3dgs}, we observe that these under-optimized baseline like 3DGS and Scaffold typically takes $25\sim40$ minutes, while those highly-optimized method like Taming-3DGS will still takes more than 10 minutes to fit a scene. In contrast, our method takes less than 10 minutes to finish the training and obtain a comparable visual quality.
\vspace{-4mm}

\paragraph{Qualitatively evaluation on 4K dataset.}
We compare our method qualitatively with 3DGS, Scaffold-GS, and Taming 3DGS in~\cref{fig:qualitative_results}. We observe that our method preserves finer details consistently.~\cref{fig:qualitative_results} shows that our method preserves fine details in regions such as grass (second and last row), pavement (fifth row) and floor (third row).
This verifies that Turbo-GS is not only fast, but also has no compromise in visual quality.

\begin{table}[!t]
    \normalsize
    \centering
    \caption{Ablations. `DR': \textbf{D}ilated \textbf{R}endering. `CG': using \textbf{C}olor \textbf{G}radient. `CA': adopt \textbf{C}onvergence-\textbf{A}ware densification. 
    In this table, both 3DGS and Scaffold-GS adopt the optimized CUDA implementation. Scaffold-GS is only trained with for 10k steps. }
    \label{tab:comparison_methods}
    \begin{tabular}{l|cccc}
        \toprule
        \textbf{Method} & \textbf{Time} & \textbf{PSNR} & \textbf{SSIM} & \textbf{LPIPS} \\
        \midrule
        3DGS          &    31m           &      26.52         &     0.791          &        0.364       \\
        \,\,+DR           &     14m          &       26.66        &     0.799          &        0.338       \\
        \,\,\,\,+CA         &       12m        &    26.58           &       0.795        &     0.347          \\
        \,\,\,\,\,\,+CG          &      16m         &       26.75        &     0.803          &   0.328            \\
        \midrule
        Scaffold-GS           &       12m        &        25.81       &     0.784          &       0.373        \\
        \,\,+DR           &       7m        &     25.58          &     0.780          &      0.375         \\
        \,\,\,\,+CA         &        8m       &       26.22        &       0.795        &       0.344        \\
        \,\,\,\,\,\,+CG          &        9m       &       26.34        &      0.797         &         0.332      \\
        \bottomrule
    \end{tabular}
    \label{tab:ablation_studies}
    \vspace{-1mm}
\end{table}

\subsection{Ablation Study}
\label{sec:ablation}

We present ablation studies on key design choices of our method in~\cref{tab:ablation_studies}. To reduce the noise from certain dataset distribution, the result of every configuration are obtained from all the 9 scenes of MipNeRF-360~\cite{mip360}, varying from indoor scenes to outdoor scenes. And for every configuration we run on two kinds of baselines, the explicit 3DGS and the explicit+implicit Scaffold-GS. Although the two backbone optimize Gaussians with different manners, e.g., in 3DGS. each Gaussian are processed independently while in Scaffold-GS local Gaussians share information through a latent representation and MLP, we could observe similar improvement trend when combining our design step by step.

Specifically, the introduction of dilated rendering acts the main role in accelerating the whole pipeline. In the 3DGS based experiments, dilated rendering accelerates the whole pipeline for $2\times$. And the convergence-aware densification serves to furtherly stabilize the training for Scaffold-GS due to avoid adding invalid gaussians. However, it will slightly damage the performance of 3DGS variants. We guess it due to the reset operation in 3DGS disturbs the training curve. Color gradient 
do not have impact on the training speed but improves the visual quality.
\section{Conclusion and Limitations}
\label{sec:conclusion}
In this work, we present Turbo-GS, an efficient method that accelerates 3DGS model fitting for high-resolution camera views. We propose two major design choices in the current optimization framework. First, we design a dilated rendering to only provide supervision on sparsely-sample pixels, which reduces the redundancy in optimization. Experiments show that this design could be readily adapted and integrated into various other 3D Gaussian pipelines. Second, to improve the quality, we introduce a convergence-aware densification to encourage spawning new Gaussians when the model is well-trained. Third, to make up for the gradient vanishing phenomenon in texture-less region, we propose a color-aided position-color densification strategy, which significantly improves the modeling of fine details and background area. This paper aims for a learning-free design, thus lacking exploration on how to integrate with a geometry foundation model. With many such models gradually changing the entire community, we will consider how to benefit from a large model in the future to achieve further acceleration.

\section{Acknowledgements}
This research was supported by NASA grant \#80NSSC23M0075, and NSF CAREER grant \#2143576. Collaboration between Brown and IISc was facilitated through Kotak Mahindra Bank's Visiting Chair Professorship for Srinath Sridhar. Venkatesh Babu Radhakrishnan was supported by Google and Kotak IISc AI-ML Centre (KIAC). Ankit Dhiman was supported by Samsung R \& D Institute India - Bangalore. 

{
    \small
    \bibliographystyle{ieeenat_fullname}
    \bibliography{main}
}

\clearpage
\maketitlesupplementary

\appendix

\tableofcontents
\addtocontents{toc}{\protect\setcounter{tocdepth}{2}}

\section{Overview}

This supplementary material is organized as follows:
\begin{itemize}
\item {\bf Implementation Details}: Additional information on the budget schedule (\cref{sec:convergence_aware_training_schedule}) and hyper-parameter settings.
\item {\bf Additional Experimental Results}: Per-scene results and an ablation study on dilated rendering.
\end{itemize}

\section{Implementation Details}

\begin{figure*}[t]
  \centering
   \includegraphics[width=1\linewidth]{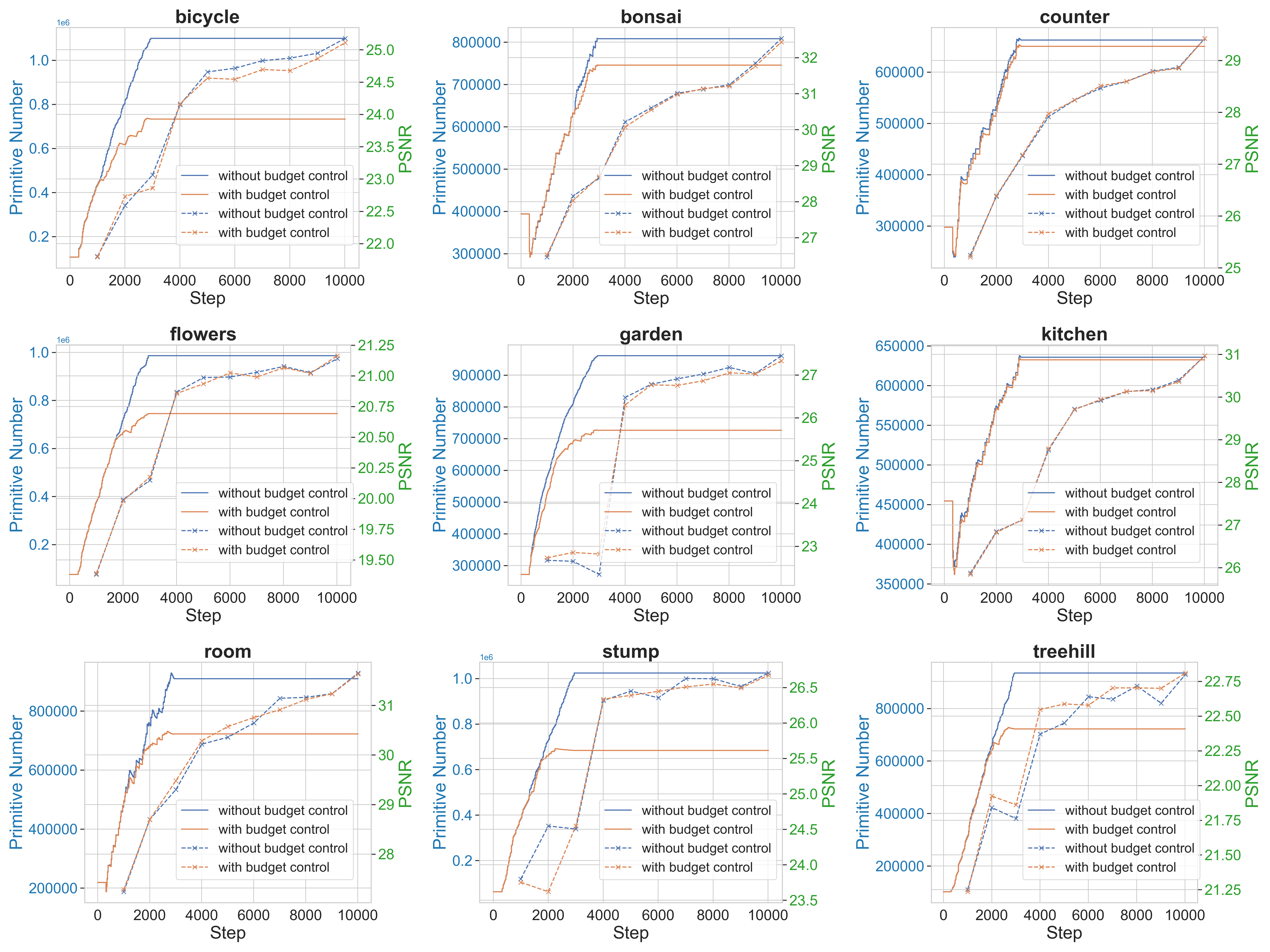}
    \vspace{-10pt}
   \caption{\textbf{Convergence.} We show ``Number of primitives vs Step'' and ``PSNR vs Step'' plots for scenes in MipNeRF-360~\cite{mip360} dataset for with and without budget control in the optimization process. The proposed budgeting strategy prevents the number of primitives from increasing uncontrollably, while maintaining the overall quality. This is evident by the comparable PSNR plots, which demonstrate that the strategy maintains the balance between computational efficiency and visual fidelity.
   }
   \vspace{-10pt}
   \label{fig:convergence}
\end{figure*}

\subsection{Convergence-aware Budget Control}

To enhance the adaptiveness of the training process based on convergence patterns, we propose a dynamic scheduling mechanism that modulates the budget for each stage according to the deviation between recent and historical trends. As illustrated in~\cref{alg:adaptive}, this approach implements two adaptations: it dynamically adjusts both the power law exponent and the final budget target. The power law exponent is tuned based on the convergence behavior in log space, while the final budget is automatically scaled up or down depending on the loss decline rate. This dual-adaptation strategy enables the scheduler to respond effectively to varying convergence dynamics while maintaining training stability.~\cref{fig:convergence} shows that with the adaptive budget control, the final number of primitives are significantly reduced while the rendering quality keeps comparable, implying that the budget control helps to add the proper number of Gaussians. And, fewer number of Gaussians denote a faster training process.

\begin{algorithm}[!t]
\SetAlgoLined
\KwIn{
   $N, M$: Numbers for initialization and final budget \\
   $steps$: Total steps \\
   $window\_size$: Window for trend analysis \\
   $t$: Current step
}
\KwOut{$B(t)$: Current scheduled value}

Initialize EMA smoother with $\alpha_{ema} = 0.1$ \\
\uIf{$len(loss\_history) \leq warmup\_steps$}{
   Use default power law with $\alpha = 1.0$
}
\Else{
   \tcp{Smoothing and Base Trend}
   Compute smoothed losses: $ema_t = \alpha_{ema} \cdot loss_t + (1-\alpha_{ema}) \cdot ema_{t-1}$ \\
   Fit in log space: $\log(ema\_losses) \sim \alpha_{base} \cdot \log(steps)$ \\
   $\alpha_{base} \gets$ Moving average of recent $\alpha_{base}$ values
   
   \tcp{Dynamic Range Adjustment}
   $rate \gets -\frac{d\log(ema\_losses)}{d\log(steps)}$ in recent window \\
   \uIf{$rate > 0.05$}{
       $M_{adaptive} \gets \min(M_{adaptive} \cdot 1.1, M \cdot 1.5)$
   }
   \ElseIf{$rate < -0.05$}{
       $M_{adaptive} \gets \max(M_{adaptive} \cdot 0.9, M \cdot 0.5)$
   }
   
   \tcp{Adaptive Power Law}
   $deviation \gets -rate - \alpha_{base}$ \\
   $\alpha \gets \alpha_{base} + 0.5 \tanh(deviation)$ \\
   Clip $\alpha$ to $[0.1, 2.0]$
}

\Return{$N + ((t^\alpha - 1)/(100^\alpha - 1)) \cdot (M_{adaptive} - N)$}

\caption{Adaptive Power Law Scheduling}
\label{alg:adaptive}
\end{algorithm}

\section{More Experiments and Results}

\subsection{Per-scene Results} Here we list the error metrics used in
our evaluation in Sec.4 across all considered methods and
scenes, as shown in~\cref{tab:psnr_comparison,tab:ssim_comparison,tab:lpips_comparison,tab:training_time_comparison}. {\bf drjohnson}-{\bf playroom}~\cite{hedman2018deep} belongs to the deep blending dataset; {\bf train}-{\bf truck} come from the Tanks and Temple~\cite{knapitsch2017tanks} dataset; {\bf bicycle}-{\bf bonsai} are from MipNeRF360~\cite{mip360}.

\begin{figure}[t]
  \centering
   \includegraphics[width=0.8\linewidth]{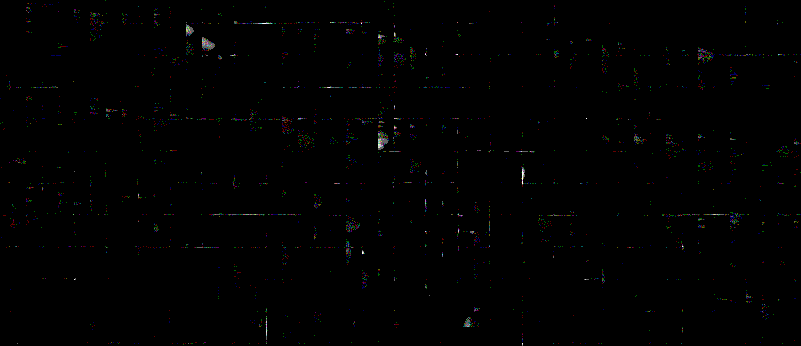}
   \caption{The error map ($\times 100$ for better view) between "render a high-resolution image" and "render 4 low resolution with different offsets then merge into a high-resolution image". The artifact locates near the tile edge. 
   } 
   \label{fig:low_to_high_error}
\end{figure}

\begin{figure}[t]
  \centering
   \includegraphics[width=1\linewidth]{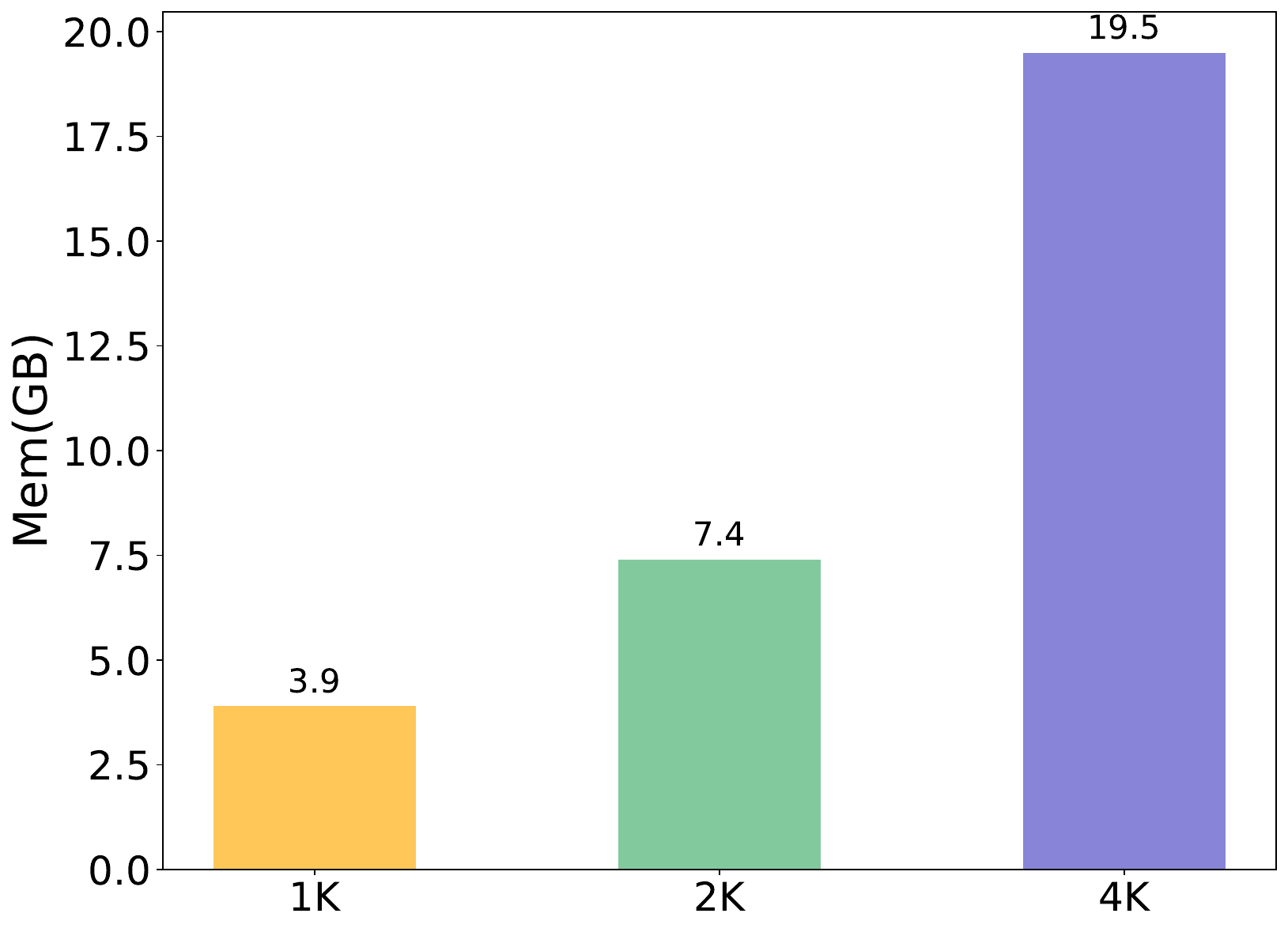}
    \vspace{-20pt}
   \caption{The memory consumption of rendering the same Gaussian checkpoint in different resolutions. We could observe a big increasement happens from 2K to 4K.
   } 
   \vspace{-10pt}
   \label{fig:memory}
\end{figure}

\subsection{Dilated Rendering}

Since the tile in low resolution corresponds to multiple tiles in the high-resolution image space, there are inevitable misalignment if mimic high-resolution rendering in low-resolution space. To analyze the distribution of this part of the error, we render a series of high-resolution and their corresponding dilated low-resolution images. The difference map is shown in~\cref{fig:low_to_high_error}.

\begin{table*}[htbp]
\centering
\caption{PSNR comparison across DeepBlending~\cite{hedman2018deep} (``drjohnson'', ``playroom''),  Tanks \& Temples~\cite{knapitsch2017tanks} (``train'', ``truck'')  and MipNeRF-360~\cite{mip360} scenes. }
\begin{adjustbox}{max width=\textwidth}
\begin{tabular}{lccccccccccccc}
\toprule
\textbf{Method} & \textbf{drjohnson} & \textbf{playroom} & \textbf{train} & \textbf{truck} & \textbf{bicycle} & \textbf{garden} & \textbf{stump} & \textbf{flowers} & \textbf{treehill} & \textbf{counter} & \textbf{kitchen} & \textbf{room} & \textbf{bonsai} \\
\midrule
3DGS~\cite{3dgs} & \cellcolor{tabthird}29.49 & 30.02 & \cellcolor{tabsecond}22.11 & 25.45 & \cellcolor{tabsecond}25.23 & 27.38 & 26.59 & 21.44 & 22.49 & 29.08 & 31.09 & 31.48 & \cellcolor{tabsecond}32.31 \\
Taming-3DGS~\cite{mallick2024taming} & 29.39 & 30.04 & \cellcolor{tabthird}22.09 & \cellcolor{tabthird}25.46 & \cellcolor{tabthird}25.22 & 27.35 & 26.62 & \cellcolor{tabthird}21.50 & 22.59 & 29.07 & 30.98 & \cellcolor{tabsecond}31.62 & 32.22 \\
Taming-3DGS*~\cite{mallick2024taming} & 29.36 & 29.94 & 21.77 & 25.29 & 25.13 & 27.24 & 26.46 & 21.43 & 22.53 & 28.95 & 31.09 & 31.18 & 32.03 \\
Mini-Splatting~\cite{fang2024mini} & 29.35 & \cellcolor{tabthird}30.46 & 21.46 & 25.06 & \cellcolor{tabthird}25.22 & 26.85 & \cellcolor{tabfirst}27.22 & \cellcolor{tabsecond}21.60 & \cellcolor{tabthird}22.68 & 28.60 & \cellcolor{tabthird}31.31 & 31.36 & 31.21 \\
EAGLES~\cite{Girish2023EAGLESEA} & 29.35 & 30.19 & 21.36 & 25.00 & 24.97 & 26.87 & 26.61 & 21.31 & 22.62 & 28.30 & 30.52 & 31.38 & 31.23 \\
ScaffoldGS~\cite{scaffoldgs} & \cellcolor{tabsecond}29.67 & \cellcolor{tabsecond}30.90 & \cellcolor{tabfirst}22.53 & \cellcolor{tabfirst}25.86 & 25.21 & \cellcolor{tabsecond}27.51 & 26.56 & 21.43 & \cellcolor{tabfirst}23.12 & \cellcolor{tabfirst}29.45 & \cellcolor{tabfirst}31.61 & \cellcolor{tabfirst}31.98 & \cellcolor{tabfirst}32.53 \\
Mip-Splatting~\cite{Yu2023MipSplattingA3} & 28.75 & 29.98 & 22.06 & \cellcolor{tabsecond}25.74 & \cellcolor{tabfirst}25.56 & \cellcolor{tabfirst}27.69 & \cellcolor{tabsecond}26.91 & \cellcolor{tabfirst}21.70 & 22.35 & \cellcolor{tabthird}29.13 & \cellcolor{tabsecond}31.59 & \cellcolor{tabthird}31.54 & \cellcolor{tabthird}32.27 \\
Turbo-Scaffold-GS & \cellcolor{tabfirst}29.85 & \cellcolor{tabfirst}31.12 & 21.75 & 25.11 & 25.01 & \cellcolor{tabthird}27.41 & \cellcolor{tabthird}26.64 & 21.18 & \cellcolor{tabsecond}22.91 & \cellcolor{tabsecond}29.29 & 30.86 & 31.50 & 32.06 \\
Turbo-3DGS & \cellcolor{tabfirst}29.85 & \cellcolor{tabfirst}31.12 & 21.75 & 25.11 & 25.01 & \cellcolor{tabthird}27.41 & \cellcolor{tabthird}26.64 & 21.18 & \cellcolor{tabsecond}22.91 & \cellcolor{tabsecond}29.29 & 30.86 & 31.50 & 32.06 \\
\bottomrule
\end{tabular}
\end{adjustbox}
\label{tab:psnr_comparison}
\end{table*}

\begin{table*}[htbp]
\centering
\caption{SSIM comparison across DeepBlending~\cite{hedman2018deep} (``drjohnson'', ``playroom''),  Tanks \& Temples~\cite{knapitsch2017tanks} (``train'', ``truck'')  and MipNeRF-360~\cite{mip360} scenes.}
\vspace{-3pt}
\begin{adjustbox}{max width=\textwidth}
\begin{tabular}{lccccccccccccc}
\toprule
\textbf{Method} & \textbf{drjohnson} & \textbf{playroom} & \textbf{train} & \textbf{truck} & \textbf{bicycle} & \textbf{garden} & \textbf{stump} & \textbf{flowers} & \textbf{treehill} & \textbf{counter} & \textbf{kitchen} & \textbf{room} & \textbf{bonsai} \\
\midrule
3DGS~\cite{3dgs} & 0.903 & 0.902 & \cellcolor{tabthird}0.818 & \cellcolor{tabthird}0.881 & \cellcolor{tabthird}0.765 & \cellcolor{tabsecond}0.864 & \cellcolor{tabthird}0.770 & \cellcolor{tabthird}0.602 & 0.633 & 0.907 & 0.925 & 0.918 & 0.940 \\
Taming-3DGS~\cite{mallick2024taming} & 0.902 & 0.904 & \cellcolor{tabthird}0.818 & \cellcolor{tabthird}0.881 & \cellcolor{tabthird}0.765 & \cellcolor{tabthird}0.863 & \cellcolor{tabthird}0.770 & 0.601 & 0.634 & 0.906 & 0.925 & 0.919 & 0.939 \\
Taming-3DGS*~\cite{mallick2024taming} & 0.903 & 0.901 & 0.812 & 0.878 & 0.751 & 0.859 & 0.764 & 0.596 & 0.630 & 0.905 & 0.923 & 0.915 & 0.939 \\
Mini-Splatting~\cite{fang2024mini} & 0.903 & \cellcolor{tabsecond}0.912 & 0.798 & 0.874 & \cellcolor{tabsecond}0.773 & 0.848 & \cellcolor{tabfirst}0.806 & \cellcolor{tabsecond}0.626 & \cellcolor{tabfirst}0.654 & 0.905 & \cellcolor{tabthird}0.926 & 0.922 & 0.939 \\
EAGLES~\cite{Girish2023EAGLESEA} & \cellcolor{tabthird}0.906 & 0.908 & 0.796 & 0.872 & 0.757 & 0.844 & \cellcolor{tabthird}0.770 & 0.589 & 0.637 & 0.897 & 0.920 & 0.917 & 0.934 \\
ScaffoldGS~\cite{scaffoldgs} & \cellcolor{tabfirst}0.907 & \cellcolor{tabsecond}0.912 & \cellcolor{tabsecond}0.822 & \cellcolor{tabsecond}0.886 & 0.760 & 0.863 & 0.766 & 0.594 & \cellcolor{tabsecond}0.643 & \cellcolor{tabthird}0.911 & \cellcolor{tabsecond}0.927 & \cellcolor{tabthird}0.924 & \cellcolor{tabfirst}0.944 \\
Mip-Splatting~\cite{Yu2023MipSplattingA3} & 0.898 & 0.908 & \cellcolor{tabfirst}0.827 & \cellcolor{tabfirst}0.893 & \cellcolor{tabfirst}0.793 & \cellcolor{tabfirst}0.878 & \cellcolor{tabsecond}0.791 & \cellcolor{tabfirst}0.640 & \cellcolor{tabthird}0.639 & \cellcolor{tabfirst}0.913 & \cellcolor{tabfirst}0.930 & \cellcolor{tabfirst}0.925 & \cellcolor{tabfirst}0.944 \\
Ours & \cellcolor{tabfirst}0.907 & \cellcolor{tabfirst}0.915 & 0.816 & \cellcolor{tabsecond}0.886 & 0.757 & 0.861 & 0.769 & 0.593 & 0.636 & \cellcolor{tabfirst}0.913 & \cellcolor{tabthird}0.926 & \cellcolor{tabfirst}0.925 & \cellcolor{tabfirst}0.944 \\
\bottomrule
\end{tabular}
\end{adjustbox}
\label{tab:ssim_comparison}
\end{table*}

\begin{table*}[htbp]
\centering
\caption{LPIPS comparison across DeepBlending~\cite{hedman2018deep} (``drjohnson'', ``playroom''),  Tanks \& Temples~\cite{knapitsch2017tanks} (``train'', ``truck'')  and MipNeRF-360~\cite{mip360} scenes.}
\begin{adjustbox}{max width=\textwidth}
\begin{tabular}{lccccccccccccc}
\toprule
\textbf{Method} & \textbf{drjohnson} & \textbf{playroom} & \textbf{train} & \textbf{truck} & \textbf{bicycle} & \textbf{garden} & \textbf{stump} & \textbf{flowers} & \textbf{treehill} & \textbf{counter} & \textbf{kitchen} & \textbf{room} & \textbf{bonsai} \\
\midrule
3DGS~\cite{3dgs} & \cellcolor{tabfirst}0.238 & \cellcolor{tabthird}0.243 & \cellcolor{tabsecond}0.198 & 0.143 & \cellcolor{tabthird}0.211 & \cellcolor{tabsecond}0.108 & \cellcolor{tabthird}0.217 & 0.339 & 0.329 & 0.201 & \cellcolor{tabthird}0.127 & 0.220 & 0.205 \\
Taming-3DGS~\cite{mallick2024taming} & \cellcolor{tabsecond}0.239 & \cellcolor{tabthird}0.243 & \cellcolor{tabthird}0.200 & 0.144 & \cellcolor{tabsecond}0.210 & \cellcolor{tabthird}0.109 & \cellcolor{tabthird}0.217 & 0.341 & 0.328 & 0.202 & \cellcolor{tabthird}0.127 & 0.219 & 0.206 \\
Taming-3DGS*~\cite{mallick2024taming} & 0.244 & 0.253 & 0.206 & 0.147 & 0.236 & 0.116 & 0.229 & 0.346 & 0.341 & 0.204 & 0.130 & 0.229 & 0.208 \\
Mini-Splatting & 0.256 & 0.249 & 0.245 & 0.160 & 0.225 & 0.150 & \cellcolor{tabsecond}0.199 & \cellcolor{tabthird}0.327 & \cellcolor{tabthird}0.313 & \cellcolor{tabthird}0.198 & 0.129 & 0.211 & \cellcolor{tabthird}0.200 \\
EAGLES~\cite{Girish2023EAGLESEA} & 0.244 & 0.253 & 0.240 & 0.166 & 0.232 & 0.146 & 0.229 & 0.361 & 0.338 & 0.217 & 0.138 & 0.226 & 0.218 \\
ScaffoldGS~\cite{scaffoldgs} & 0.252 & 0.253 & 0.206 & \cellcolor{tabthird}0.142 & 0.227 & 0.118 & 0.236 & 0.347 & 0.319 & 0.200 & \cellcolor{tabthird}0.127 & \cellcolor{tabthird}0.210 & 0.203 \\
Mip-Splatting~\cite{Yu2023MipSplattingA3} & 0.243 & \cellcolor{tabsecond}0.235 & \cellcolor{tabfirst}0.189 & \cellcolor{tabfirst}0.123 & \cellcolor{tabfirst}0.167 & \cellcolor{tabfirst}0.094 & \cellcolor{tabfirst}0.188 & \cellcolor{tabfirst}0.274 & \cellcolor{tabfirst}0.274 & \cellcolor{tabsecond}0.187 & \cellcolor{tabfirst}0.119 & \cellcolor{tabsecond}0.202 & \cellcolor{tabfirst}0.188 \\
Ours & \cellcolor{tabthird}0.241 & \cellcolor{tabfirst}0.224 & \cellcolor{tabthird}0.200 & \cellcolor{tabsecond}0.132 & 0.224 & 0.116 & 0.221 & \cellcolor{tabsecond}0.320 & \cellcolor{tabsecond}0.293 & \cellcolor{tabfirst}0.185 & \cellcolor{tabsecond}0.126 & \cellcolor{tabfirst}0.196 & \cellcolor{tabsecond}0.191 \\
\bottomrule
\end{tabular}
\end{adjustbox}
\label{tab:lpips_comparison}
\end{table*}

\begin{table*}[htbp]
\centering
\caption{Training time (in minutes) comparison across DeepBlending~\cite{hedman2018deep} (``drjohnson'', ``playroom''),  Tanks \& Temples~\cite{knapitsch2017tanks} (``train'', ``truck'')  and MipNeRF-360~\cite{mip360} scenes.}
\begin{adjustbox}{max width=\textwidth}
\begin{tabular}{lccccccccccccc}
\toprule
\textbf{Method} & \textbf{drjohnson} & \textbf{playroom} & \textbf{train} & \textbf{truck} & \textbf{bicycle} & \textbf{garden} & \textbf{stump} & \textbf{flowers} & \textbf{treehill} & \textbf{counter} & \textbf{kitchen} & \textbf{room} & \textbf{bonsai} \\
\midrule
3DGS & 35.04 & 26.13 & 16.25 & 19.27 & 35.60 & 34.64 & 29.31 & 24.94 & 27.21 & 29.27 & 34.97 & 30.32 & 24.52 \\
Taming-3DGS & \cellcolor{tabthird}15.71 & \cellcolor{tabthird}11.02 & \cellcolor{tabthird}8.11 & \cellcolor{tabthird}11.74 & \cellcolor{tabthird}22.58 & \cellcolor{tabthird}22.31 & 18.88 & \cellcolor{tabthird}14.65 & \cellcolor{tabthird}15.37 & \cellcolor{tabthird}11.44 & \cellcolor{tabthird}15.51 & \cellcolor{tabthird}11.35 & \cellcolor{tabthird}9.92 \\
Taming-3DGS* & \cellcolor{tabsecond}9.40 & \cellcolor{tabsecond}6.81 & \cellcolor{tabsecond}6.22 & \cellcolor{tabsecond}7.51 & \cellcolor{tabsecond}11.95 & \cellcolor{tabsecond}12.72 & \cellcolor{tabsecond}9.17 & \cellcolor{tabsecond}8.35 & \cellcolor{tabsecond}8.53 & \cellcolor{tabsecond}9.37 & \cellcolor{tabsecond}14.52 & \cellcolor{tabsecond}8.28 & \cellcolor{tabsecond}7.60 \\
Mini-Splatting & 19.23 & 16.89 & 13.94 & 13.87 & 16.72 & 18.65 & \cellcolor{tabthird}16.55 & 18.07 & 18.14 & 29.22 & 28.83 & 24.67 & 24.47 \\
EAGLES & 28.48 & 19.45 & 12.34 & 13.18 & 23.70 & 21.88 & 22.58 & 17.36 & 20.85 & 22.18 & 27.81 & 24.08 & 19.57 \\
ScaffoldGS & 17.64 & 16.87 & 15.02 & 14.68 & 22.41 & 23.99 & 18.08 & 21.05 & 21.14 & 27.32 & 32.84 & 23.41 & 24.42 \\
Mip-Splatting & 42.21 & 32.31 & 18.72 & 28.12 & 59.24 & 53.88 & 43.57 & 38.28 & 40.68 & 32.87 & 39.80 & 35.08 & 29.84 \\
Ours & \cellcolor{tabfirst}4.04 & \cellcolor{tabfirst}3.98 & \cellcolor{tabfirst}4.80 & \cellcolor{tabfirst}4.71 & \cellcolor{tabfirst}5.34 & \cellcolor{tabfirst}7.70 & \cellcolor{tabfirst}5.43 & \cellcolor{tabfirst}5.64 & \cellcolor{tabfirst}6.05 & \cellcolor{tabfirst}8.60 & \cellcolor{tabfirst}9.77 & \cellcolor{tabfirst}6.36 & \cellcolor{tabfirst}6.93 \\
\bottomrule
\end{tabular}
\end{adjustbox}
\label{tab:training_time_comparison}
\end{table*}

\begin{table}[!t]
\centering
\caption{Sensitivity to $\lambda$}
\begin{adjustbox}{width=0.7\linewidth}
\begin{tabular}{c|cccc}
\toprule
\textbf{$\lambda = $} & \textbf{0.2} & \textbf{0.7} & \textbf{1.0} & \textbf{0.5} (Ours) \\ 
\midrule
\textbf{PSNR $\uparrow$} & 27.406 & 27.346 & 27.370 & \textbf{27.423} \\
\textbf{SSIM $\uparrow$} & 0.813 & 0.812 & 0.813 &  \textbf{0.814} \\  
\textbf{LPIPS $\downarrow$} & \textbf{0.221} & \textbf{0.221} & \textbf{0.221} &  \textbf{0.221} \\
\bottomrule
\end{tabular}
\end{adjustbox}
\label{tab:sensitivity_to_lambda}
\end{table}

\subsection{Sensitivity to $\lambda$}
We evaluate $\lambda$ described in~\cref{eq:lambda_equation} for $\lambda \in \{0.2, 0.5, 0.7, 1.0\}$ in~\cref{tab:sensitivity_to_lambda}. We observe optimal PSNR at $\lambda = 0.5$, which is used for all our experiments.

\begin{table}[!t]
\centering
\caption{Comparison on different budget schedules.}
\begin{adjustbox}{width=0.9\linewidth}
\begin{tabular}{c|cccccc}
\toprule
  &  \textbf{linear} & \textbf{beizer} & \textbf{cosine} & \textbf{exp} & \textbf{log}   & \textbf{Ours} \\
  \midrule
 \textbf{PSNR $\uparrow$} & 27.093 & 27.055 & 27.098 & 27.159 & 27.126 &  \textbf{27.423}\\
 \textbf{SSIM $\uparrow$} & 0.798 & 0.797 & 0.800 & 0.801 & 0.801 & \textbf{0.813} \\
 \textbf{LPIPS $\downarrow$} & 0.253 & 0.253 & 0.250 & 0.249 & 0.249 & \textbf{0.221} \\

\bottomrule
\end{tabular}
\end{adjustbox}
\label{tab:supp_diff_budget_schedules}
\end{table}

\subsection{Different Budget Schedules}
We evaluate alternative budget schedules for~\cref{eq:budget_schedule} in~\cref{tab:supp_diff_budget_schedules}. We observe that our adaptive convergence-aware schedule consistently outperforms other schedules. 

\begin{table}[!t]
\centering
\caption{Impact of position-appearance based densification}
\begin{adjustbox}{width=0.5\linewidth}
\begin{tabular}{c|cc}
\toprule

 & \makecell{Only \\ Position} &  \makecell{Position + \\ Color (Ours)}\\ 
\midrule
\textbf{PSNR $\uparrow$} &  27.285 & \textbf{27.423}\\
\textbf{SSIM $\uparrow$} &  0.804 &  \textbf{0.814} \\  
\textbf{LPIPS $\downarrow$}  & 0.238 &  \textbf{0.221}\\
\end{tabular}
\end{adjustbox}
\label{tab:supp_position_apperance_densification_impact}
\end{table}

\subsection{Impact of position-appearance based densification}
We visualize the magnitude of color and position gradients via alpha-blending in~\cref{fig:gradient_viz} and observe that color gradients are significantly 
more active in fine-textured (background) regions than position gradients. Impact of using both gradients during densification improve results qualitatively (\cref{fig:qualitative_results}) and is quantitatively validated in~\cref{tab:supp_position_apperance_densification_impact}, where the ``Only Position'' variant yields degraded performance.

\begin{figure}[!t]
    \centering
    \includegraphics[width=\linewidth]{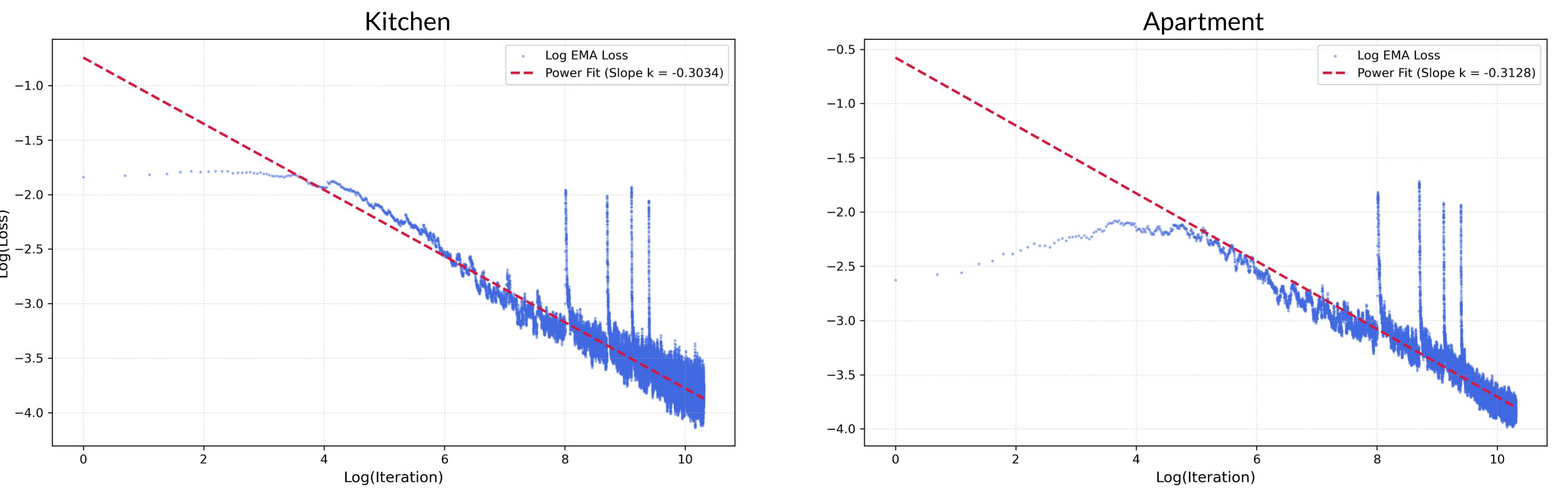}
    \caption{Loss analysis with power function fitting. \textbf{(Zoom-in)}}
    \label{fig:supp_power_law_4k}
\end{figure}

\subsection{Analysis of power function fitting on 4K scenes}
Our convergence-aware training schedule is motivated by the observation that the $\log(\text{loss})$ vs. $\log(\text{iteration})$ curve exhibits stable linear behavior after initial stage, indicating a power-law function (\cref{fig:powerfit}). We further validate this hypothesis in~\cref{fig:supp_power_law_4k} using two 4K scenes from the EyefulTower~\cite{eyefultower}  dataset, which confirms that the log-linear relationship holds after the initial optimization phase.

\subsection{Memory Consumption} In the main paper we show the timebreakdown for different resolution rendering. Here we plot the memory consumption for different resolutions in~\cref{fig:memory}.

\begin{table}[t]
\centering
\caption{Original Image resolutions of scenes in the MipNeRF-360 dataset.}
\begin{tabular}{lcc}
\toprule
\textbf{Scene} & \textbf{Width} & \textbf{Height} \\
\midrule
bicycle   & 4946 & 3286 \\
flower    & 5025 & 3312 \\
garden    & 5187 & 3361 \\
stump     & 4978 & 3300 \\
treehill  & 5068 & 3326 \\
\bottomrule
\end{tabular}
\label{tab:mipnerf360_resolutions}
\end{table}

\subsection{Comparison on MipNeRF-360 4K dataset.}
MipNeRF-360~\cite{mip360} contains nine scenes but only has five scenes with $\geq4K$ resolution. We describe these five scenes in~\cref{tab:mipnerf360_resolutions}. All these scenes are outdoor scenes. 
~\cref{tab:mipnerf4k_supp} compares our method and recent baselines on this challenging high resolution dataset.

\begin{table}[!t]
\caption{Quantitative comparison with Gaussian Splatting baselines for 4K scenes. We compare PSNR, SSIM, and LPIPS for quality. For resource efficiency, we report training time, memory and, where applicable and peak number (Peak \#G) of Gaussians. '-accel' means accelerated with optimized CUDA implementation.}
\vspace{-5pt}
\label{tab:supp_results_4k}
\begin{adjustbox}{width=\linewidth}
\begin{tabular}{l|c|c|c|c|c|c|}
\multicolumn{1}{c}{} & \multicolumn{6}{c|}{MipNeRF-360 4K~\cite{mip360}}  \\ \hline   
                             & \multicolumn{1}{c|}{SSIM$\uparrow$}           & \multicolumn{1}{c|}{PSNR$\uparrow$}           & \multicolumn{1}{c|}{LPIPS$\downarrow$}          & \multicolumn{1}{c|}{\begin{tabular}[c]{@{}c@{}}Train\\ time $\downarrow$\end{tabular}} & \multicolumn{1}{c|}{\begin{tabular}[c]{@{}c@{}}Memory $\downarrow$\end{tabular}} & \multicolumn{1}{c|}{\begin{tabular}[c]{@{}c@{}}Peak \\ \#G $\downarrow$\end{tabular}} \\
 \hline  
3DGS~\cite{3dgs}     & 0.797 & 26.75 & 0.410 & 113~m & 412~MB & \textbf{1.74~M}  \\
3DGS-accel~\cite{3dgs}     &0.791 & 26.52 & 0.364 & 40~m & 383~MB & 1.62~M  \\
SpeedySplat~\cite{speedygs}     & 0.678 & 23.23 & 0.467 & 97~m & \textbf{65.6~MB} & 1.79~M  \\
Turbo-3DGS (ours)     & \textbf{0.805} & \textbf{26.80}& \textbf{0.327}& \textbf{24~m} & 680~MB & 2.80~M \\
\hline
\end{tabular}
\end{adjustbox}
\vspace{-10pt}
\label{tab:mipnerf4k_supp}
\end{table}

\end{document}